\documentclass[journal]{IEEEtran}

\usepackage{amsmath,amssymb,graphicx}
\usepackage{multirow}
\usepackage{epstopdf}
\usepackage{rotating}
\usepackage{mathrsfs}
\usepackage{amssymb}

\usepackage[caption=false]{subfig}
\usepackage{algorithm}
\usepackage{algorithmic}
\usepackage{xcolor}
\usepackage{makecell}
\usepackage{array}
\usepackage{hyperref}

\ifCLASSINFOpdf

\else

\fi
\begin{document}
	\title{Towards Unsupervised Deep Image Enhancement with Generative Adversarial Network}

	\author{Zhangkai Ni,~\IEEEmembership{Graduate Student Member,~IEEE}, Wenhan Yang,~\IEEEmembership{Member,~IEEE}, \\ Shiqi Wang,~\IEEEmembership{Member,~IEEE}, Lin Ma,~\IEEEmembership{Member,~IEEE}, and Sam Kwong,~\IEEEmembership{Fellow,~IEEE}
    \thanks{This work was supported in part by the Natural Science Foundation of China under Grant 61772344, and Grant 61672443, in part by the Hong Kong Research Grants Council (RGC) General Research Funds under Grant 9042816 (CityU 11209819), and under Grant 9042957 (CityU 11203220), in part by the Hong Kong Research Grants Council (RGC) Early Career Scheme under Grant 9048122 (CityU 21211018), and in part by the Key Project of Science and Technology Innovation 2030 supported by the Ministry of Science and Technology of China under Grant 2018AAA0101301.
    \emph{(Corresponding authors: Shiqi Wang and Sam Kwong)}.}
    \thanks{Zhangkai Ni, Wenhan Yang, and Shiqi Wang are with the Department of Computer Science, City University of Hong Kong, Hong Kong 999077 (e-mail: eezkni@gmail.com; yangwenhan@pku.edu.cn; shiqwang@cityu.edu.hk).}
    \thanks{Lin Ma is with the Meituan-Dianping Group, Beijing 100102, China (e-mail: forest.linma@gmail.com).}
    \thanks{Sam Kwong is with the Department of Computer Science, City University of Hong Kong, Hong Kong 999077, and also with the City University of Hong Kong Shenzhen Research Institute, Shenzhen 518057, China (e-mail: cssamk@cityu.edu.hk).}
    }
	
	\maketitle
	
	\begin{abstract}
    Improving the aesthetic quality of images is challenging and eager for the public. To address this problem, most existing algorithms are based on~\textit{supervised} learning methods to learn an automatic photo enhancer for~\textit{paired} data, which consists of low-quality photos and corresponding expert-retouched versions. However, the style and characteristics of photos retouched by experts may not meet the needs or preferences of general users. In this paper, we present an~\textit{unsupervised} image enhancement generative adversarial network (UEGAN), which learns the corresponding~\textit{image-to-image} mapping from a set of images with desired characteristics in an~\textit{unsupervised} manner, rather than learning on a large number of paired images. The proposed model is based on~\textit{single} deep GAN which embeds the~\textit{modulation} and \textit{attention} mechanisms to capture richer global and local features. Based on the proposed model, we introduce two losses to deal with the unsupervised image enhancement: (1)~\textit{fidelity loss}, which is defined as a $\ell2$ regularization in the feature domain of a pre-trained VGG network to ensure the content between the enhanced image and the input image is the same, and (2)~\textit{quality loss} that is formulated as a relativistic hinge adversarial loss to endow the input image the desired characteristics. Both quantitative and qualitative results show that the proposed model effectively improves the aesthetic quality of images. Our code is available at: \url{https://github.com/eezkni/UEGAN}.

	\end{abstract}
	

	%
	\IEEEpeerreviewmaketitle
	
	\section{Introduction}
	\label{sec:intro}
	
	\IEEEPARstart{W}{ith} the rapid development of mobile Internet, smart electronic devices, and social networks, it is becoming more and more popular to record and upload the wonderful lives of people through social media and online sharing communities. However, due to the high cost of high-quality hardware devices and the lack of professional photography skills, the aesthetic quality of photos taken by the general public is often unsatisfactory. Professional image-editing is expensive, and it is hard to provide such services in an automated manner as aesthetic feelings and preferences are usually a personal issue. Therefore, the \textit{automatic image enhancement techniques} providing the~\textit{user-oriented} image beautification are preferred.

Compared with high-quality images, low-quality images usually suffer from multiple degradations in visual quality, such as poor colors, low contrast, and intensive noises~\textit{et al}. Therefore, the image enhancement process needs to address this degradation with a series of enhancement operations, such as contrast enhancement, color correction, and details adjustment~\textit{et al}. The earliest conventional image enhancement approaches mainly focused on contrast enhancement of low-quality image~\cite{arici2009histogram, thomas2011histogram, lee2013contrast}. The most common histogram adjustment transfers the luminance histogram of a low-quality image to a given distribution (may be provided by other reference images) to stretch the contrast of the low-quality image. According to the transformation scope, this kind of method can be further classified into two categories:~\textit{global} histogram equalization (GHE)~\cite{thomas2011histogram, coltuc2006exact} and~\textit{local} histogram equalization (LHE)~\cite{lee2013contrast, abdullah2007dynamic}. The former uses a single histogram transformation function to adjust all pixels of the entire image. It may lead to improper enhancement results in some local regions, such as under-exposure, over-exposure, color distortion, \textit{et al}. To address this issue, the LHE derives the content adaptive transform functions based on the statistical information in local region and applies these transforms locally. However, the LHE is computationally complex and not always powerful because the extracted transformation depends on the dominating information in the local region. Therefore, they are also easy to generate visually unsatisfactory texture details, dull or over-saturated color.
	
	For the past few years, deep convolutional neural networks (CNN) have made significant progress in low-level vision tasks~\cite{gharbi2017deep, ignatov2017dslr, yan2016automatic}. In order to improve the modeling capacity and adaptivity, deep learning-based models are built to introduce the excellent expressive power of deep networks to facilitate automatic image enhancement with the knowledge of big data. Ignatov~\textit{et al.}~\cite{ignatov2017dslr} designed an end-to-end deep learning network that improves photos from mobile devices to the quality of digital single-lens reflex (DSLR) photos. Ren~\textit{et al.}~\cite{ren2019low} present a hybrid loss to optimize the framework from three aspects (\textit{i.e.,} color, texture, and content) to produce more visually pleasing results. Inspired by bilateral grid processing, Gharbi~\textit{et al.}~\cite{gharbi2017deep} made real-time image enhancement possible, which dynamically generates the image transformation based on local and global information. To deal with low-light image enhancement, Wang~\textit{et al.}~\cite{wang2019underexposed} established a large-scale under-exposed image dataset and learned an image-to-illumination mapping based on the Retinex model to enhance extremely low-light images.

\begin{figure}[t]
\centering
\subfloat[][]{\includegraphics[width=0.155\textwidth]{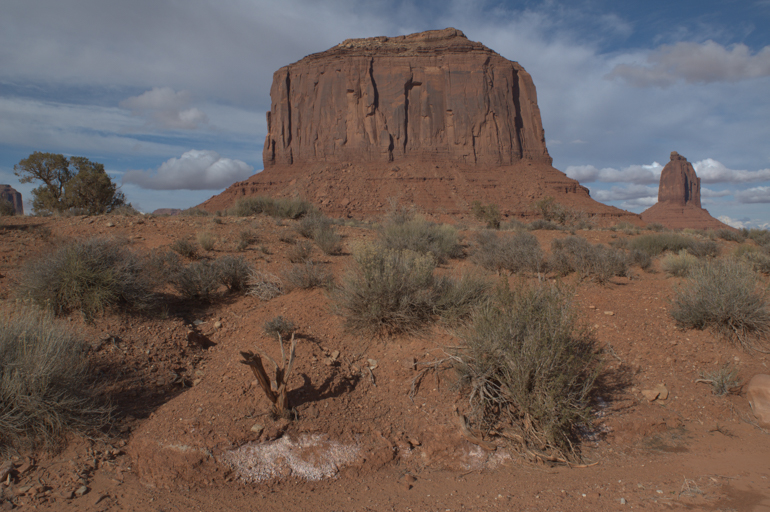}}\hfil%
\subfloat[][]{\includegraphics[width=0.155\textwidth]{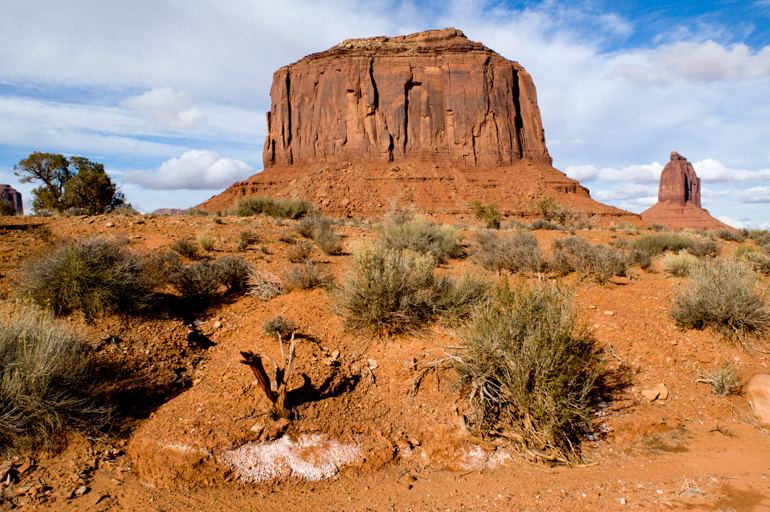}}\hfil%
\subfloat[][]{\includegraphics[width=0.155\textwidth]{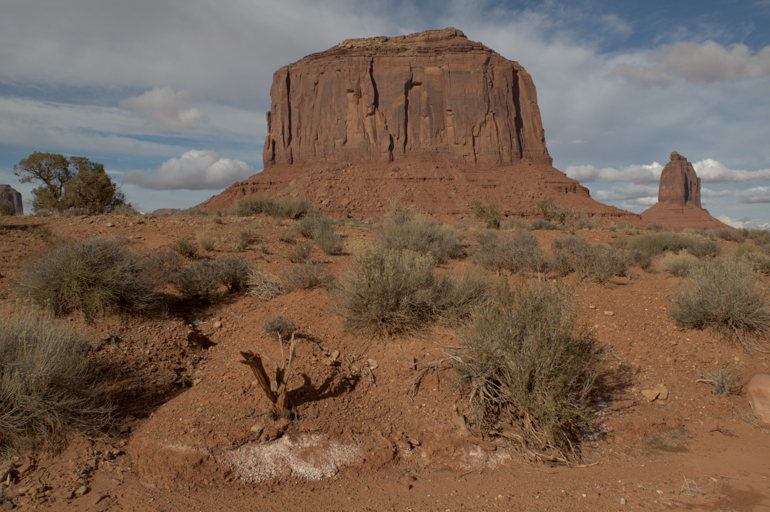}} \\
\vspace{-0.6em}
\subfloat[][]{\includegraphics[width=0.155\textwidth]{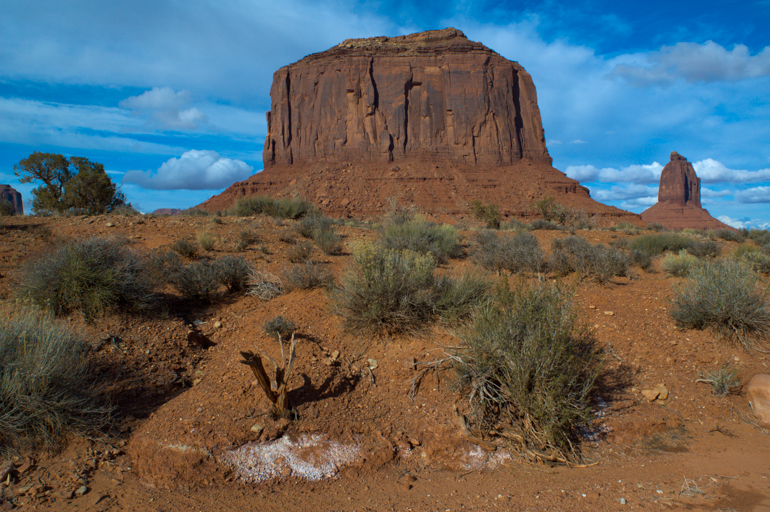}}\hfil%
\subfloat[][]{\includegraphics[width=0.155\textwidth]{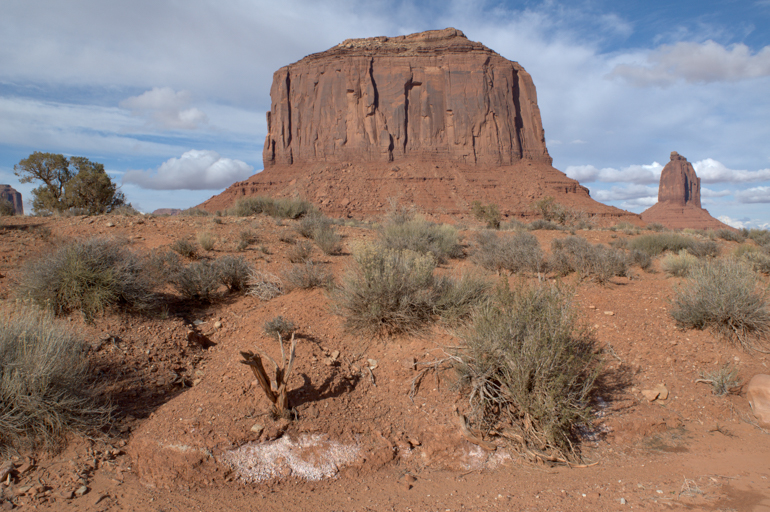}}\hfil%
\subfloat[][]{\includegraphics[width=0.155\textwidth]{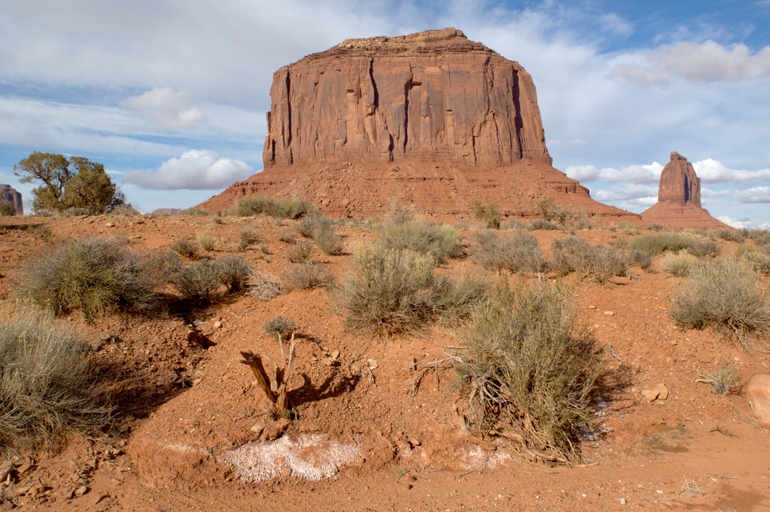}}  \\
\caption{An illustration of different expert-retouched versions of a low-quality photo in MIT-Adobe FiveK dataset~\cite{bychkovsky2011learning}. (a) is the low-quality photo and (b) to (f) are different high-quality counterparts retouched by different experts. The obvious perceptual differences exist among different high-quality versions.}
\label{fig:fivekexample}
\end{figure}

	However, these methods follow the route of~\emph{fully supervised} learning relying on large-scale datasets with~\textit{paired low}/\textit{high}-quality images. First, paired data is usually expensive, and sometimes it takes a lot of effort and resources to build the dataset by professional photographers. Second, the judgment of image quality is usually closely related to the personality, aesthetics, taste, and experience of a person. ``There are a thousand Hamlets in a thousand people's eyes.'' In other words, everyone has his/her different attitude towards the quality of the photography. To demonstrate this, a typical low-quality photo in MIT-Adobe FiveK Dataset~\cite{bychkovsky2011learning} and its corresponding five high-quality versions retouched by five different experts in photo beautification, are shown in Fig.~\ref{fig:fivekexample}, respectively. It can be observed that the images processed by one expert are very different from the image retouched by another expert. Consequently, it is impractical to create a large-scale dataset with paired low and high-quality images to meets the preference of everyone. On the contrary, a more feasible way is to express the personal preferences of a user by providing a set of image collections that he/she loves. Therefore, an urgent demand is needed to build an enhancement model to learn the enhancement mapping from the low-quality dataset to a high-quality one even without the specific paired images. In this way, we can get rid of the burden of creating one-to-one paired data and rely only on the~\textit{target} dataset with the desired characteristics preferred by someone.

Benefit from the development of generative adversarial learning~\cite{zhu2017unpaired, chen2018deep, lin2019coco} and reinforcement learning (RL)~\cite{hu2018exposure}, some works make attempts to handle the image enhancement tasks only with the help of unpaired data. The milestone work of transferring image style between unpaired data is CycleGAN~\cite{zhu2017unpaired}. It employs two generators and two discriminators and uses cycle consistency loss to achieve visually impressive results. Chen~\textit{et al.}~\cite{chen2018deep} proposes to construct a bi-directional GAN with three improvements to transfer low-quality images into corresponding high-quality ones, and the experimental results show that this model is significantly better than CycleGAN. Hu~\textit{et al.}~\cite{hu2018exposure} design the first RL-based framework to train an effective photo post-processing model. Jiang~\textit{et al.}~\cite{jiang2019enlightengan} carry out the first study on the task of low-light enhancement with an unsupervised framework. The method applies a self-regularized attention generator and dual discriminators to guide the generator globally and locally.

	Rather than utilizing a~\textit{cyclic} generative adversarial network (GAN) to learn bi-directional mappings between the low-quality photos and high-quality ones, we build a~\textit{unidirectional} GAN to address the image~\textit{aesthetic} quality enhancement task, called the unsupervised image enhancement GAN (UEGAN). Inspired by the properties as mentioned above, our network consists of a joint global and local generator and a multi-scale discriminator with effective constraints.
	1) The generator consists of an encoder and decoder with a~\textit{global attention module} and a~\textit{modulation module} embedded, which adjusts the features at different scales locally and globally. The~\textit{multi-scale} discriminator also inspects the results at different levels of granularity and guides the generator to produce better results to obtain global consistency and finer details.
	2) To keep the content invariance, a~\textit{fidelity loss} is introduced to regularize the consistency between the input content and resulting content.
    3) The global features extracted from the entire image reveal high-level information such as lighting conditions and color distributions. To capture these properties, the~\textit{global attention module} is designed to adjust pixels according to the information of a local neighborhood to meet both local adaptivity and global consistency.
	4) For preventing over-enhancement, an~\textit{identity loss} is introduced to constrain the consistency between the enhanced result of the input high-quality image and the input one. This benefits controlling the enhancement procedure to be more quality-free and thus prevents over-enhancement.
	The main contributions of this work are summarized as follows:
	\begin{itemize}
		\item
        We design a single GAN framework that gets rid of the needs of~\textit{paired} training data for image~\textit{aesthetic} quality enhancement. To the best of our knowledge, this is the first trial to employ a~\textit{unidirectional} GAN framework to apply~\textit{unsupervised} learning to enhance the aesthetic quality of images (instead of low-light image enhancement).
        \item
        We propose a~\textit{global attention module} and a~\textit{modulation module} to construct the joint global and local generator to capture global features and adaptively adjust the local features. Together with the proposed multi-scale discriminator to inspect the quality of the generated results at different scales, well-enhanced results in perception and aesthetics are produced with both global consistency and finer details.
		\item
        We propose to jointly use~\textit{quality loss},~\textit{fidelity loss}, and~\textit{identity loss} to train our model to make it towards extracting quality-free features and controlling the enhancement procedure to be more robust to the quality change. Thus, our method can obtain more reasonable results and prevent over-enhancement. Extensive experimental results on various datasets demonstrate the superiority of the proposed model quantificationally and qualitatively.
	\end{itemize}
	
	The remaining of this paper is organized as follows. In Section~\ref{sec:related}, the related work is succinctly described. In Section~\ref{sec:proposed}, the proposed unsupervised image aesthetic quality enhancement model is presented in detail. In Section~\ref{sec:results}, extensive experimental results of the proposed are reported. In Section~\ref{sec:discussion}, the ablation studies and analysis are presented. Finally, Section~\ref{sec:conclusion} draws the conclusion.
	

	\section{Related Work}
	\label{sec:related}
	
	\subsection{Traditional Image Enhancement}
	\label{ssec:traditional}
    Extensive research has been conducted over the past few decades to improve the quality of photos. Most existing conventional image enhancement algorithms aim to stretch contrast and improve sharpness. The following three types of approaches are the most representative:~\textit{histogram adjustment}, ~\textit{unsharp masking}, and~\textit{Retinex-based approaches}. These approaches are succinctly described as follows.
	
	\textit{1) Histogram Adjustment}. Based on the basic idea of mapping the luminance histogram to a specific distribution, many methods estimate the mapping function based on the statistical information of the entire image~\cite{thomas2011histogram, coltuc2006exact, ibrahim2007brightness}, while the details usually tend to be over-enhanced due to the dominance of some high-frequency information. Instead of estimating a single mapping function for the entire image, other approaches dynamically adjust the histogram based on~\textit{local} statistical information~\cite{lee2013contrast, abdullah2007dynamic, stark2000adaptive}. However, higher computational complexity limits the applicability of this method.
	
	\textit{2) Unsharp Masking}. Unsharp masking (UM) aims to improve image sharpness~\cite{ye2018blurriness}. The framework of the UM approach can be summarized into the following two phases: First, the input image is decomposed into a~\textit{base layer} and a~\textit{detail layer} by applying a low or high pass filter. Second, all pixels in the detail layer are scaled by a~\textit{single} global weighting factor, or different pixels are adaptively scaled by pixel-wise weighting factors, and then added back to the base layer to obtain an enhanced version. Various works have been proposed to improve the performance of UM from two aspects: 1) design a more reasonable layer decomposition method to decouple different frequency bands~\cite{mitra1991new, he2012guided}; and 2) propose a better estimation algorithm for the adjustment scaling factor~\cite{ye2018blurriness, polesel2000image}.

	\textit{3) Retinex-Based Approaches}. Many researchers are working on Retinex-based image enhancement due to clear physical meaning. The basic assumption of the Retinex model is that the observed photo can be decomposed into reflection and illumination~\cite{li2018structure}. The enhanced image depends on the decomposed layer, \textit{i.e.}, illumination and reflectance layers. Therefore, the Retinex-based model is usually approached as an illumination estimation problem~\cite{wang2013naturalness, guo2017lime, ying2017new}. However, such approaches might generate unnatural results due to the ambiguity and difficulty in accurately estimating the illumination and reflection map.
	
	\begin{figure*}[t]
		\centering
        \centerline{\includegraphics[width=1\linewidth]{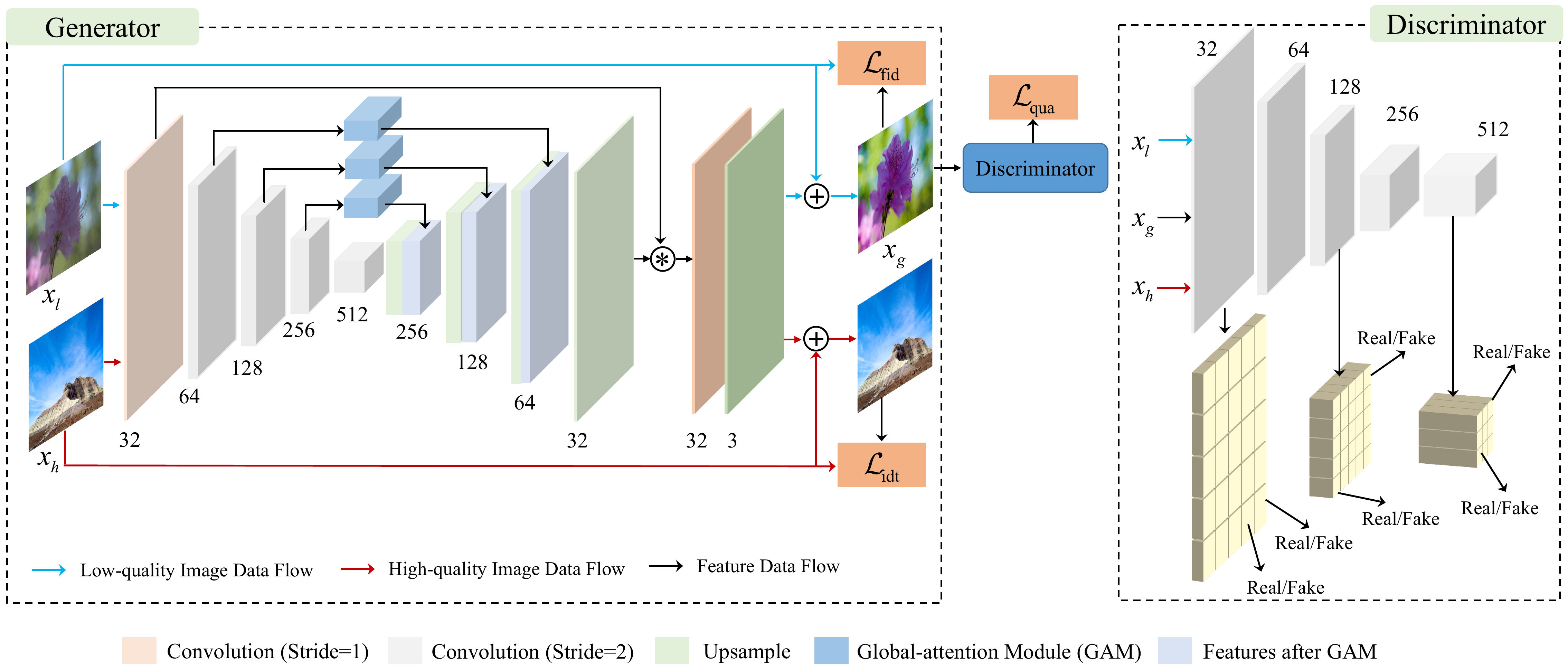}}
		\caption{The framework of the proposed UEGAN for image enhancement. The blue, red, and black lines indicate the low-quality images data flow, high-quality images data flow, and features data flow, respectively. The generator only inputs low-quality or high-quality images at a time. }
		\label{fig:framework}
	\end{figure*}
	
	\subsection{Learning-based Image Enhancement}
	\label{ssec:learning}
	
	\textit{1) Supervised Learning Approaches}. Given the explosive growth of CNN, image enhancement models based on learning methods have emerged in large numbers with impressive results. Yan~\textit{et al.}~\cite{yan2016automatic} took the first step in exploring the use of CNN for photo editing. Ignatov~\textit{et al.}~\cite{ignatov2017dslr} build a large-scale DSLR
Photo Enhancement Dataset (\textit{i.e.}, DPED), which consists of 6K photos captured simultaneously by a DSLR camera and three smartphones, respectively. With the paired data, it is easy to learn a mapping function between the low-quality photos captured by smartphones and the high-quality photos captured by the professional DSLR camera. Ren~\textit{et al.}~\cite{ren2019low} proposed a hybrid framework to address the low-light enhancement problem by jointly considering the content and structure. However, the promising performance of these models is inseparable from the premise of a large number of pairs of degraded images and corresponding high-quality counterparts.
	
	\textit{2) Unsupervised Learning Approaches}. Different from super-resolution, deraining, and denoising, the high-quality images are usually already present, and their low-quality versions can be easily generated by degrading them. In most cases, the image enhancement requires generating high-quality counterparts from low-quality images if need paired low-/high-quality during the training phase. High-quality photos are usually obtained by experts using professional photo editing programs (\textit{i.e.}, Adobe Photoshop and Lightroom) to retouch low-quality photos. This is expensive, time-consuming, and the editing style might depend heavily on the expert rather than the real users. In order to get rid of paired training data, a few works attempted to address the image enhancement issue with unsupervised learning. Inspired by the well-known CycleGAN~\cite{zhu2017unpaired}, Chen~\textit{et al.}~\cite{chen2018deep} designed a dual GAN model to learn a bi-directional mapping between the source domain and target domain. Specifically, the learned transformation from the source domain to the target domain is first used to generate the high-quality image, and then the inverse mapping from the target domain to the source domain is learned to translate the generated high-quality image back to the source domain. The cycle consistency loss is constrained to enforce the closeness between the input low-quality photos and those generated by the reverse translation. The cycle consistency works well if both bi-directional generators provide an ideal mapping between the two domains. However, the instability of GAN increases training difficulty and risk to local minima when the cycle consistency is applied.

	
	\section{Proposed unsupervised GAN for image enhancement}
	\label{sec:proposed}
    \subsection{Motivations and Objectives}
	\label{ssec:motivations}

    We observe that professional photographer usually follows these instincts when performing image editing:
    \begin{itemize}	
    	\item \textit{Combination of global transformation and local adjustment}. The content and intrinsic semantic information should be kept the same between the low-quality and retouched versions. The expert might first perform a global transformation based on the overall lighting conditions (\textit{e.g.}, well-exposure or under/over-exposure) and tone (\textit{e.g.}, cool or warm colors) in the scenes. The local corrections then make finer adjustments based on the joint consideration of both global information and local content.
    	\item \textit{Over-enhancement prevention}. The trade-off between~\textit{fidelity} and~\textit{quality} is crucial. \textit{Over-enhancement} donates the visual effects caused by excessively enhancing the properties of images related to the aesthetic feeling, such as very warm colors, high contrast, and over-exposure, \textit{etc}. However, this can also make the results to deviate from fidelity and produce unnatural results. That is, a good automatic photo enhancer should be aware of over-enhancement while producing good visual effects.
    \end{itemize}
	
    Base on the observations mentioned above, we are dedicated to learning an~\textit{image-to-image} mapping function $\mathscr{F}$ to generate the high-quality counterpart $x_g$ of a given low-quality photo $x_l$, which can be modeled as follows,
    \begin{equation}
    \label{equ:i2i}
    x_g = \mathscr{F}(x_l).
    \end{equation}
    One critical issue in image enhancement tasks is how to define quality as well as high quality. Any user can easily provide a collection of images expressing their personal preferences without explicitly stating the quality he/she loves. Therefore, rather than defining $\mathscr{F}$ as various clearly defined rules, it is better to formulate it as a process of transforming low-quality image distribution under the guidance of the desired high-quality image distribution. This promotes us to learn a~\textit{user-oriented} photo enhancer based on~\textit{unpaired} data in an unsupervised manner. Based on this consideration, we make efforts in utilizing the set-level supervision of GAN to achieve our goals through adversarial learning.

	\subsection{Network Architecture}
\label{ssec:generator}

\begin{figure}[t]
  \centering
  \centerline{\includegraphics[width=0.80\linewidth]{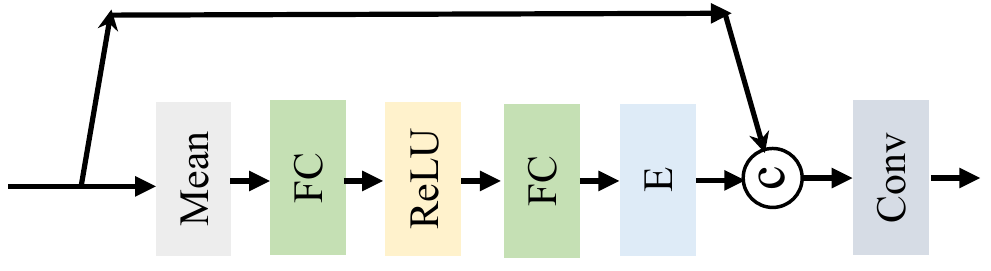}}
  \caption{The structure of the global attention module, where $E$ and $C$ denote the expanding and concatenation operations, respectively.
  }
\label{fig:ga}
\end{figure}

\begin{figure*}[t]
\centering
\subfloat[][Input]{\includegraphics[width=0.245\textwidth]{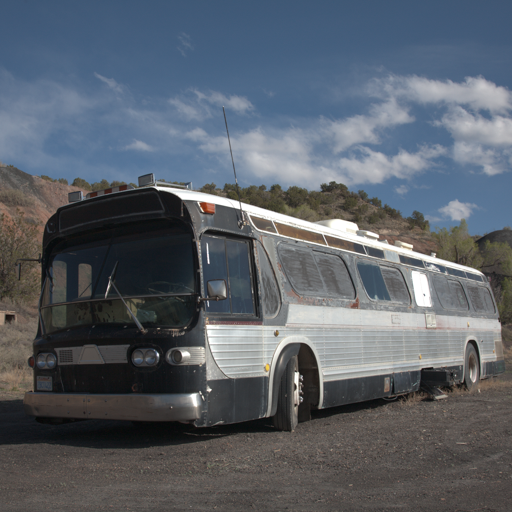}}\hfil%
\subfloat[][CycleGAN~\cite{zhu2017unpaired}]{\includegraphics[width=0.245\textwidth]{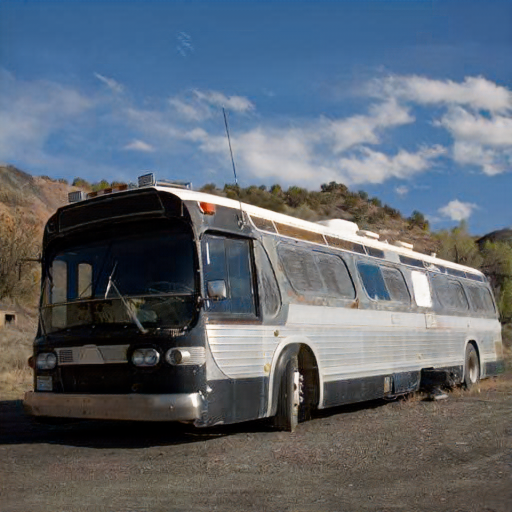}}\hfil%
\subfloat[][Exposure~\cite{hu2018exposure}]{\includegraphics[width=0.245\textwidth]{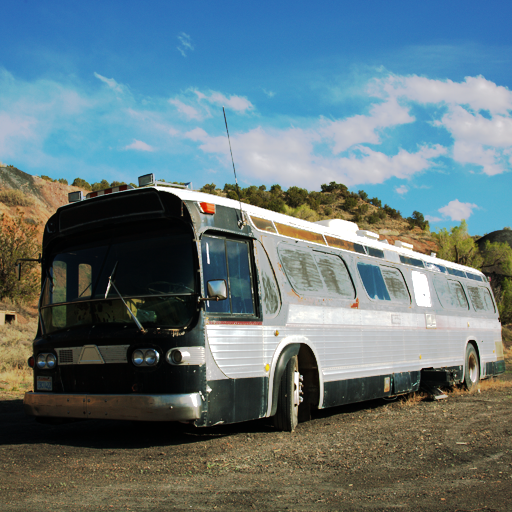}}\hfil%
\subfloat[][EnlightenGAN~\cite{jiang2019enlightengan}]{\includegraphics[width=0.245\textwidth]{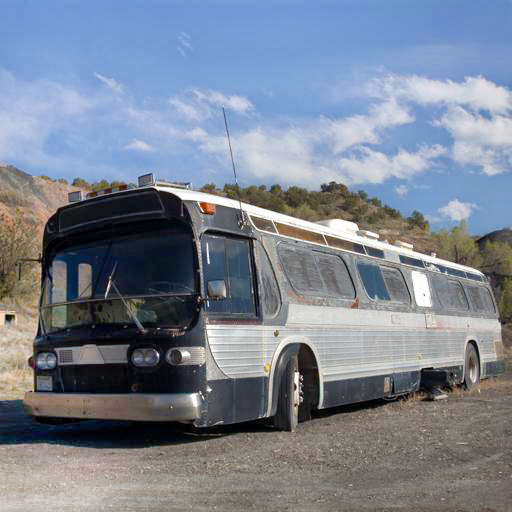}}\\
\vspace{-0.8em}
\subfloat[][DPE~\cite{chen2018deep}]{\includegraphics[width=0.245\textwidth]{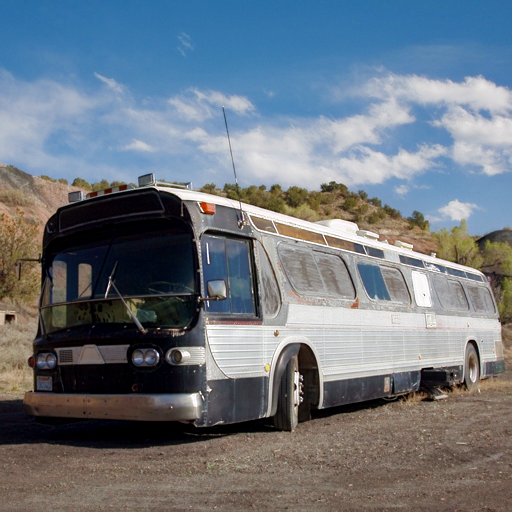}}\hfil%
\subfloat[][Ours (FiveK)]{\includegraphics[width=0.245\textwidth]{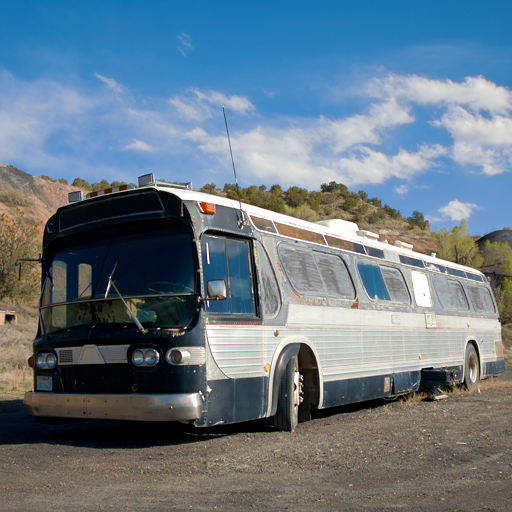}}\hfil%
\subfloat[][Ours (Flickr)]{\includegraphics[width=0.245\textwidth]{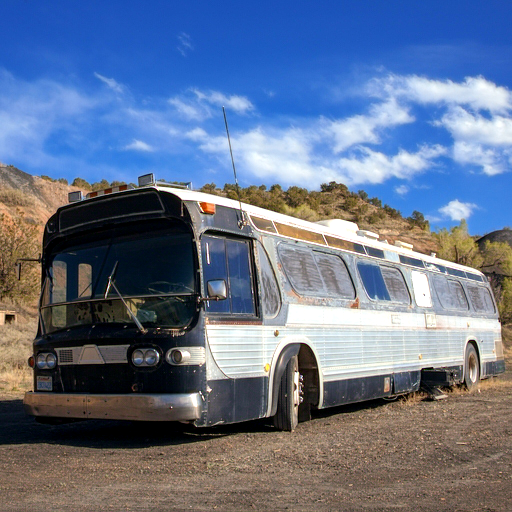}}\hfil%
\subfloat[][Expert-retouched]{\includegraphics[width=0.245\textwidth]{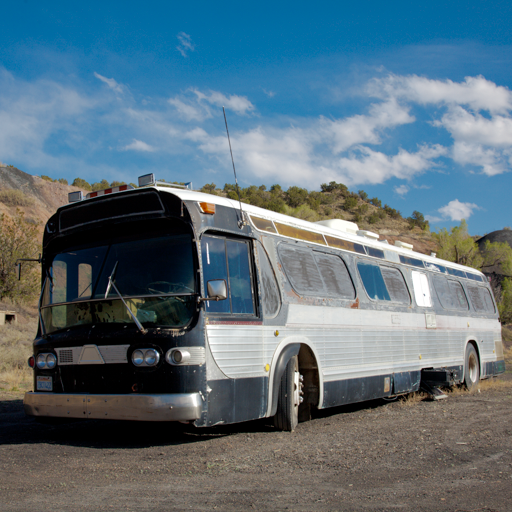}}\\
\caption{Visual quality comparison with state-of-the-art methods (\textit{i.e.}, CycleGAN, DPE, EnlightenGAN, and Exposure) on a test image from the MIT-Adobe FiveK~\cite{bychkovsky2011learning} dataset.}
\label{fig:comparison1}
\end{figure*}

\textit{1) Joint Global and Local Generator}:
The generator plays a crucial role in our proposed UEGAN as it directly affects the quality of the final generated photos. The expert might first perform a global transform based on the overall lighting conditions or tone in the scenes. Therefore, the global features act as an image prior to guiding the generation and adjusting the local features. Based on this observation, we first propose a~\textit{global attention module} (GAM) to exploit the global attention of local features. Each channel of feature maps is extracted from the local neighborhood by the convolution layer. The focus of global attention is the `holistic' understanding of each channel. In order to model the global attention of the intermediate features $z\in \mathbb{R}^{C\times H\times W}$, our proposed method can be summarized as the following three steps as shown in Fig.~\ref{fig:ga}: 1) extracting global statistics information $f^{\text{m}}_{\text{pool}}(\cdot)$ of each channel via Eqn.~(\ref{equ:mean}); 2) digging the inter-channel relationship $\rho$ using the extracted $g_{\text{mean}}$ via the multi-layer perceptron $f_{\text{FC}}(\cdot)$ in Eqn.~\eqref{equ:fc}; 3) fusing global and local features via Eqn.~(\ref{equ:fusing}).

\begin{equation}
\label{equ:mean}
   g_{\text{mean}} = f_{\text{pool}}^{\text{m}}(z),
\end{equation}
\begin{equation}
\label{equ:fc}
\rho = f_{\text{FC}}(g_{\text{mean}}),
\end{equation}
\begin{equation}
\label{equ:fusing}
\hat{z} = \text{Conv}(C(E(\rho), z)),
\end{equation}
where $f_{\text{pool}}^{\text{m}}(\cdot)$ means the average pooling operation, $f_{\text{FC}}(\cdot)$ is two fully-connected layers, $E(\cdot)$ represents expanding the spatial dimension of $\rho$ to that of $z$, $C(\cdot)$ is the concatenation operation, and $Conv(\cdot)$ is a convolution layer.

Fig.~\ref{fig:framework} shows the proposed~\textit{modulation module} (MM) in the joint global and local generator. In particular, we use skip connections between encoder and decoder at different scales locally and globally to prevent the information loss caused by resolution change. Unlike traditional U-Net~\cite{ronneberger2015u}, the features of the encoder are concatenated to those of the symmetric decoder at each stage (\textit{i.e.}, four stages in our model). Our proposed modulation module learns to generate two branches of features and then merge them together with the multiplication operation. In our model, to further reuse the features, the learned modulation layer multiples the features of the first stage of the encoder and those of the penultimate layer by element-wise multiplication. Learning global features and feature modulation can effectively enhance the visual effect of the resulting image. The global features can also guide to penalize some low-quality features that might lead to visual artifacts or poorly reconstructed details. Complex image processing can be approximated by a set of simple local smoothing curves~\cite{chen2016bilateral}, the proposed joint global and local generator $G$ is more capable than traditional U-Net for learning complex mappings from low-quality images to high-quality ones.

\textit{2) Multi-scale Discriminator}:
In order to distinguish between real high-quality image and generated ``pseudo" high-quality image, the discriminator requires a large receptive field to capture the global characteristics. This directly leads to the need for deeper networks or larger convolution kernels. The last layer of the discriminator usually captures the information from a larger region of the image and can guide the generator to produce the image with better global consistency. However, the intermediate layer of the discriminator with a smaller receptive field can force the generator to pay more attention to finer details. Based on this observation, as shown in Fig.~\ref{fig:framework}, we propose a multi-scale discriminator $D$ that uses multi-scale features to guide the generator to produce images with both global consistency and finer details.

\begin{figure*}[t]
\centering
\subfloat[][Input]{\includegraphics[width=0.245\textwidth]{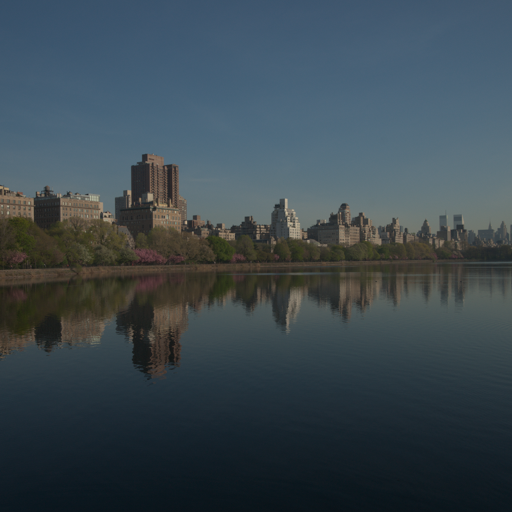}}\hfil%
\subfloat[][CycleGAN~\cite{zhu2017unpaired}]{\includegraphics[width=0.245\textwidth]{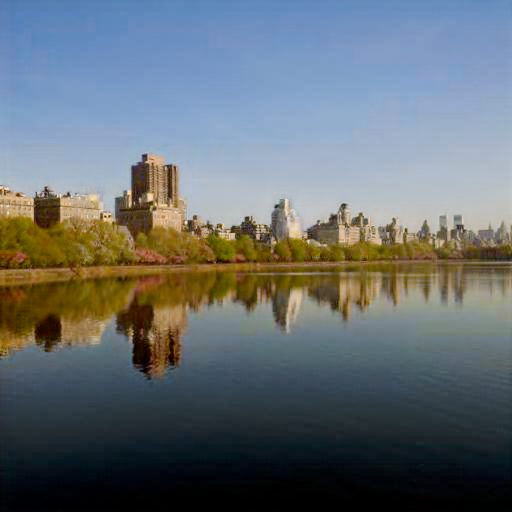}}\hfil%
\subfloat[][Exposure~\cite{hu2018exposure}]{\includegraphics[width=0.245\textwidth]{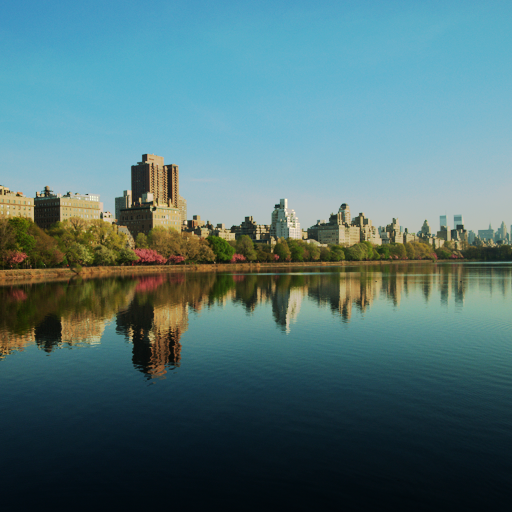}}\hfil%
\subfloat[][EnlightenGAN~\cite{jiang2019enlightengan}]{\includegraphics[width=0.245\textwidth]{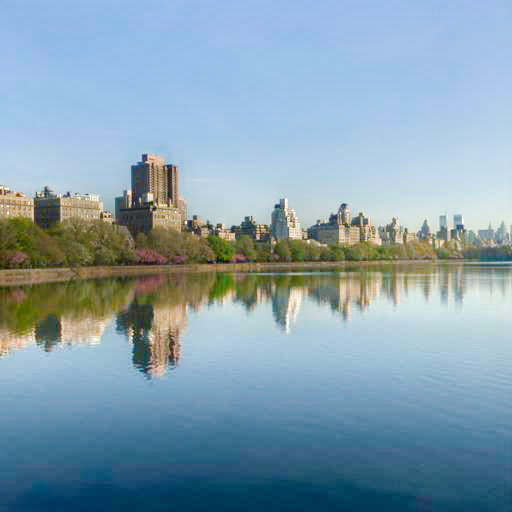}}\\
\vspace{-0.8em}
\subfloat[][DPE~\cite{chen2018deep}]{\includegraphics[width=0.245\textwidth]{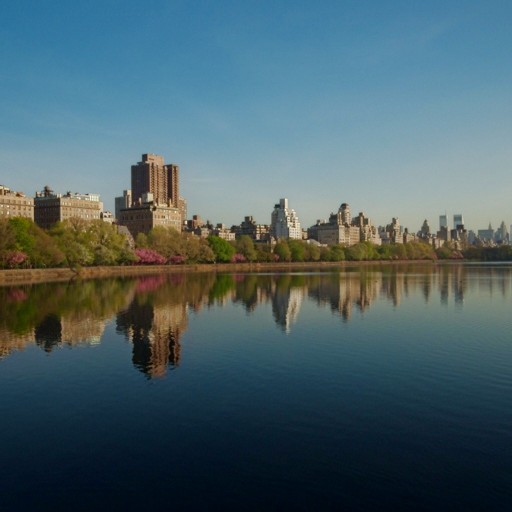}}\hfil%
\subfloat[][Ours (FiveK)]{\includegraphics[width=0.245\textwidth]{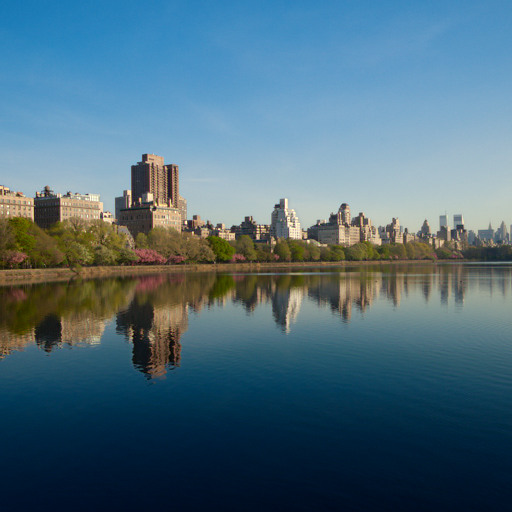}}\hfil%
\subfloat[][Ours (Flickr)]{\includegraphics[width=0.245\textwidth]{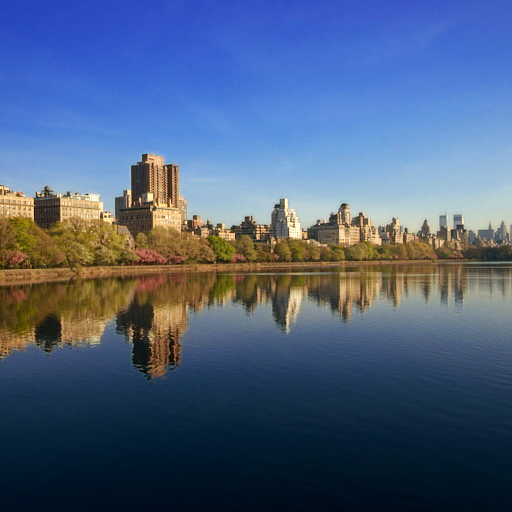}}\hfil%
\subfloat[][Expert-retouched]{\includegraphics[width=0.245\textwidth]{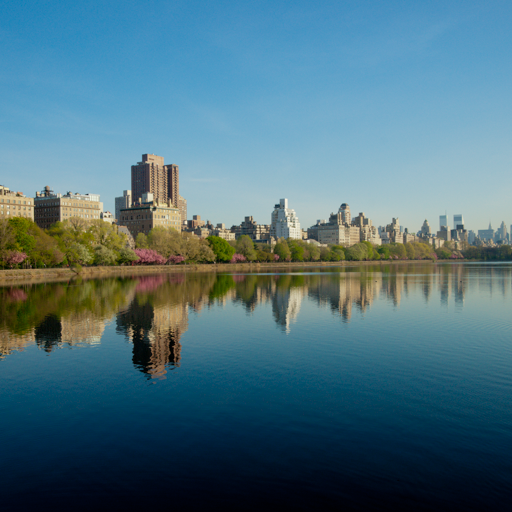}}\\
\caption{Visual quality comparison with state-of-the-art methods (\textit{i.e.}, CycleGAN, DPE, EnlightenGAN, and Exposure) on a test image from the MIT-Adobe FiveK~\cite{bychkovsky2011learning} dataset.}
\label{fig:comparison2}
\end{figure*}

\begin{figure*}[htp]
\centering
\subfloat[][Input]{\includegraphics[width=0.245\textwidth]{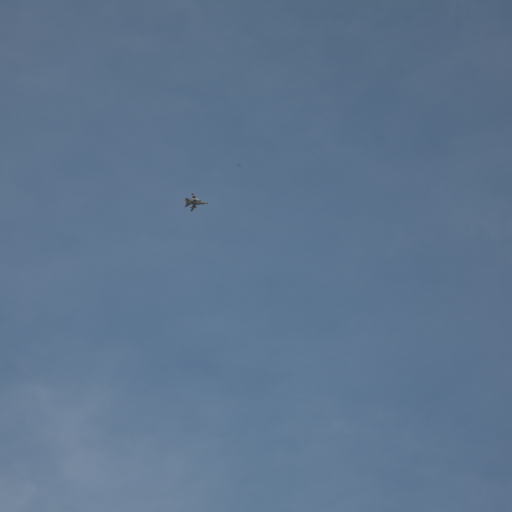}}\hfil%
\subfloat[][CycleGAN~\cite{zhu2017unpaired}]{\includegraphics[width=0.245\textwidth]{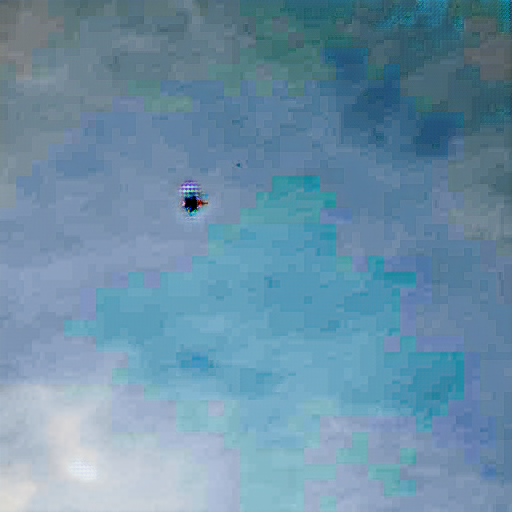}}\hfil%
\subfloat[][Exposure~\cite{hu2018exposure}]{\includegraphics[width=0.245\textwidth]{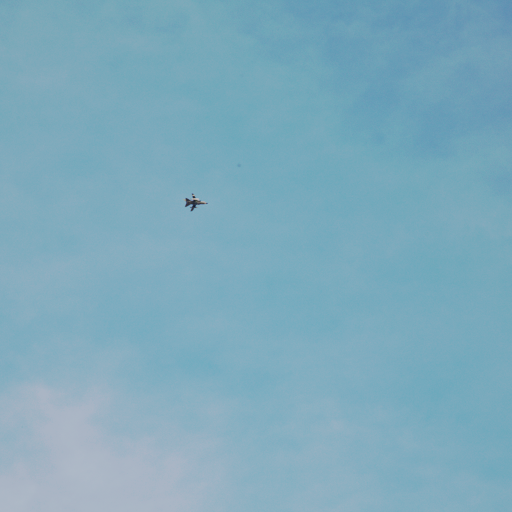}}\hfil%
\subfloat[][EnlightenGAN~\cite{jiang2019enlightengan}]{\includegraphics[width=0.245\textwidth]{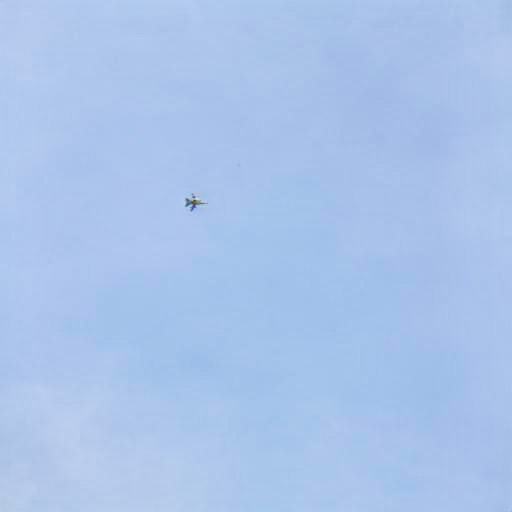}}\\
\vspace{-0.8em}
\subfloat[][DPE~\cite{chen2018deep}]{\includegraphics[width=0.245\textwidth]{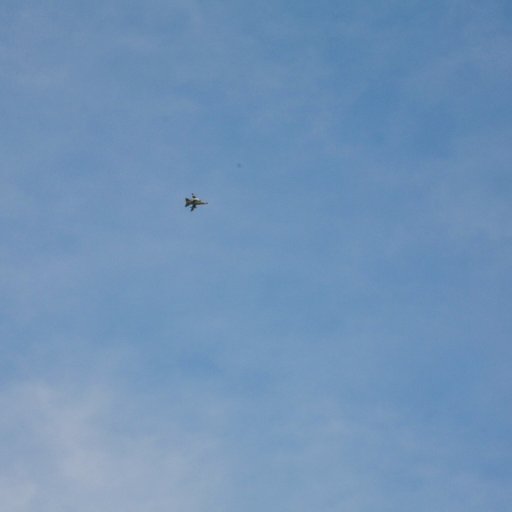}}\hfil%
\subfloat[][Ours (FiveK)]{\includegraphics[width=0.245\textwidth]{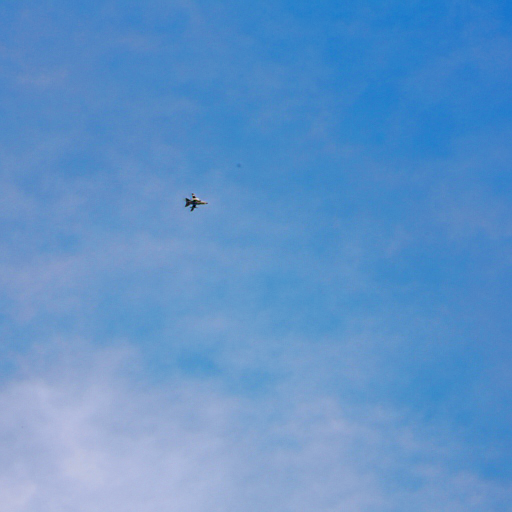}}\hfil%
\subfloat[][Ours (Flickr)]{\includegraphics[width=0.245\textwidth]{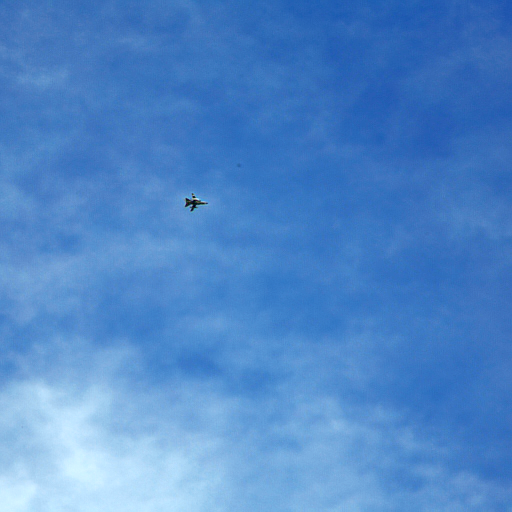}}\hfil%
\subfloat[][Expert-retouched]{\includegraphics[width=0.245\textwidth]{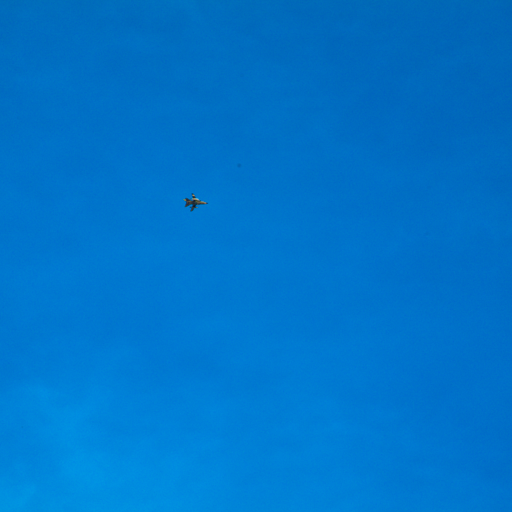}}\\
\caption{Visual quality comparison with state-of-the-art methods (\textit{i.e.}, CycleGAN, DPE, EnlightenGAN, and Exposure) on a test image from the MIT-Adobe FiveK~\cite{bychkovsky2011learning} dataset.}
\label{fig:comparison3}
\end{figure*}

\subsection{Loss Function}
\label{ssec:loss}

\textit{1) Quality Loss:} We use quality loss to adapt the distribution of enhanced results to that of high-quality images. The quality loss guides the generator to produce more visually pleasing results. In the previous GAN frameworks, the discriminator aims at distinguishing between real samples and the generated ones. However, we observe that simply applying the discriminator $D$ to separate generated images and real high-quality images is not enough to obtain a good generator that transfers low-quality images into high-quality ones. The reason might be lies in that the quality ambiguity between low/high-quality images, some images in the low-quality image set are better than those in the high-quality image set. To address this issue, we also train the discriminator to distinguish between real low-quality images and real high-quality images as shown in Fig.~\ref{fig:framework}.

Specifically, our proposed discriminator is based on the recently proposed relativistic discriminator structure~\cite{jolicoeur2018relativistic}, which not only assesses the probability that the real data (\textit{i.e.}, real high-quality image) is more authentic than the fake data (\textit{i.e.}, generated high-quality image or real low-quality image), but also guides the generator to produce high-quality images more realistic than real high-quality images. In addition, we employ an improved form of the relativistic discriminator, Relativistic average HingeGAN (RaHingeGAN)~\cite{jolicoeur2018relativistic, zhang2019self} as follows:
\begin{equation}
\begin{split}
\label{equ:dadvloss}
\mathcal{L}^{D}_{\text{}} =
\mathbb{E}_{x_l\sim P_{l}}\left[\text{max}(0, 1 + (D(x_l) - E_{x_h\sim P_{h}}D(x_h)))\right]\\
+ \mathbb{E}_{x_h\sim P_{h}}\left[\text{max}(0, 1 - (D(x_h)-E_{x_l\sim P_{l}}D(x_l)))\right] \\
+ \mathbb{E}_{x_g\sim P_{g}}\left[\text{max}(0, 1 + (D(x_g) - E_{x_h\sim P_{h}}D(x_h)))\right] \\
+ \mathbb{E}_{x_h\sim P_{h}}\left[\text{max}(0, 1 - (D(x_h)-E_{x_g\sim P_{g}}D(x_g)))\right],
\end{split}
\end{equation}
\begin{equation}
\begin{split}
\label{equ:gadvloss}
\mathcal{L}^{G}_{\text{qua}} =
\mathbb{E}_{x_h\sim P_{h}}\left[\text{max}(0, 1 + (D(x_h) - E_{x_g\sim P_{g}}D(x_g)))\right] \\
+ \mathbb{E}_{x_g\sim P_{g}}\left[\text{max}(0, 1 - (D(x_g)-E_{x_h\sim P_{h}}D(x_h)))\right],
\end{split}
\end{equation}
where $x_l$, $x_h$, and $x_g$ denote the real low-quality image, real high-quality image, and generated high-quality image, respectively.

\textit{2) Fidelity Loss:} Since we train our model for image enhancement in an unsupervised manner, the quality loss itself might not ensure that the generated image has similar content to that of the input low-quality image. The simplest way is to measure the distance between the input and output images in the pixel domain. However, we cannot employ this strategy because the generated high-quality image is typically different from the input low-quality image in the pixel domain due to contrast stretching and color rendering. Therefore, we use fidelity loss to constrain the training of the generator, so as to achieve the purpose of generated high-quality images and inputting low-quality images with similar content. The fidelity loss is defined as the $\ell_2$ norm between the feature maps of the input low-quality image and those of the generated high-quality images extracted by the pre-trained VGG network~\cite{simonyan2014very} as follows:
\begin{equation}
\label{equ:fidelityloss}
\mathcal{L}_{\text{fid}} = \sum\nolimits_{j=1}^J\{\mathbb{E}_{x_l\sim P_{l}}\left[\left\|\phi_j(x_l) - \phi_j(G(x_l))\right\|_2\right] \},
\end{equation}
where $\phi_j(\cdot)$ indicates the process of extracting the feature maps obtained by the $j^{th}$ layer of the VGG network and $J$ is the total number of layers used. Specifically, the $Relu\_1\_1$, $Relu\_2\_1$, $Relu\_3\_1$, $Relu\_4\_1$, and $Relu\_5\_1$ layers of VGG-19 network are adopted in this work.

\textit{3) Identity Loss:} The identity loss is defined as $\ell_1$ distance between the input high-quality image and the corresponding output of the generator $G$ as follows:
\begin{equation}
\label{equ:identityloss}
\mathcal{L}_{\text{idt}} = \mathbb{E}_{x_h\sim P_{h}}\left[\left\|x_h - G(x_h)\right\|_1\right].
\end{equation}
The identity loss is calculated based on high-quality input images. Therefore, if the color distribution and contrast of the input image meet the characteristics of the high-quality image set, the identity loss intends to encourage preservation of the color distributions and contrast between the input and output. It ensures that the generator should make almost no changes to the image in content, contrast, and color during the image enhancement process. As a result, the identity loss makes it possible to simultaneously maintain the content, color rendering, and contrast of the input high-quality image.

\textit{4) Total Loss:} By jointly considering~\textit{quality loss},~\textit{fidelity loss}, and~\textit{identity loss}, our final loss is defined as the weighted sum of these losses which as follows:
\begin{equation}
\label{equ:perceptualloss}
\mathcal{L}_{\text{total}} =\lambda_{\text{qua}}\mathcal{L}^G_{\text{qua}} + \lambda_{\text{fid}}\mathcal{L}_{\text{fid}} + \lambda_{\text{idt}}\mathcal{L}_{\text{idt}} ,
\end{equation}
where $\lambda_{\text{qua}}$, $\lambda_{\text{fid}}$, and $\lambda_{\text{idt}}$ are weighting parameters to balance the relative importance of $\mathcal{L}^G_{\text{qua}}$, $\mathcal{L}_{\text{fid}}$ and $\mathcal{L}_{\text{idt}}$.

\section{Experimental Results}
\label{sec:results}

\subsection{Dataset}
\label{ssec:dataset}

\textit{1) MIT-Adobe FiveK Dataset:} This dataset was constructed by Bychkovsky~\textit{et al.}~\cite{bychkovsky2011learning} for the image enhancement task, where high-quality images are generated by experts retouching. It consists of 5000 raw photos and 25,000 retouched photos generated from those raw photos by five experienced photographers. Therefore, this dataset includes five subsets, each with 5,000 raw and corresponding retouched photo pairs. Following the works in~\cite{chen2018deep, hu2018exposure}, we select the retouched photos generated by photographer C as the target photos (\textit{i.e.}, ground truth) since the user rates this subset best. In order to generate unpaired training data, the subset is randomly divided into three partitions: 1) the first partition has 2,250 raw photos as low-quality input; 2) the second partition consists of retouched version of another 2,250 raw photos and served as the desired high-quality photos; 3) the last partition is the remaining 500 raw photos used for validation (100 images) and testing (400 images). These three parts have no overlaps with each other.

\textit{2) Flickr Dataset:} In addition to training on the photographer results of the MIT-Adobe FiveK Dataset, we also collected a high-quality image collection from Flickr for unpaired training. These images are crawled from the Flickr images tagged with ``High Dynamic Range" to ensure relatively consistent quality and then manually selected by the authors. Finally, we select 2,000 images as the desired high-quality labels.

\subsection{Implementation Details}
\label{ssec:implementation}

We built our network in Pytorch and train it for 150 epochs on an NVidia GeForce RTX 2080 Ti GPU with a mini-batch size of 10. The entire network is optimized from scratch using Adam optimizer~\cite{kingma2014adam} with a learning rate of 0.0001. The leaning rate is fixed at the first 75 epochs and then linearly decays to zero in the next 75 epochs. For the MIT-Adobe FiveK Dataset, we use Lightroom to decode the images into the png format and resize the long side of the images to 512 resolution. For data augmentation, we randomly cropped 256$\times$256 patches from images.

In the MIT-Adobe FiveK Dataset, we set the hyper-parameters $\lambda_{\text{qua}}$, $\lambda_{\text{fid}}$, and $\lambda_{\text{idt}}$ as 0.05, 1, and 0.1, respectively as empirically these values provide the best performance in quantitative and qualitative performance. When coming to the Flickr Dataset, the hyper-parameters $\lambda_{\text{qua}}$, $\lambda_{\text{fid}}$, and $\lambda_{\text{idt}}$ are also empirically set as 0.05, 1 and 0.1.

\subsection{Evaluation Metrics}
\label{ssec:metrics}

The most commonly-used full-reference image quality assessment metrics (\textit{i.e.}, PSNR and SSIM) focus only on~\textit{signal fidelity} but may not accurately reflect aesthetic and perceptual quality. Although the evaluation of aesthetic quality is challenging, we still have the tool to measure the enhancement quality to an extent with the quantitative evaluation. To this end, the NIMA~\cite{talebi2018nima} score is used to quantify the aesthetic quality. The NIMA is an effective CNN-based image aesthetic quality assessment method trained on the large-scale aesthetic dataset AVA~\cite{murray2012ava}. It is predicts the distribution of human opinion scores rather than the mean opinion scores (\textit{i.e}, MOS). Therefore, we use PSNR, SSIM, and NIMA to compare our proposed method with the state-of-the-art methods at the pixel level, structural level, and aesthetics level, where the first two metrics are performed in terms of the similarity between the enhanced results and the corresponding expert-retouched (\textit{i.e.}, ground truth). In general, higher PSNR, SSIM and NIMA values correspond to reasonably better results.

\begin{table}[t]
\footnotesize
\renewcommand{\arraystretch}{1.5}
\caption{Quantitative comparison between our proposed method and state-of-the-art methods on MIT-Adobe FiveK Dataset~\cite{bychkovsky2011learning}}
\label{tab:quantitative}
\tabcolsep0.4cm
\centering
\begin{tabular}{l|c|c|c}
\hline
\hline
Method  &  PSNR  &    SSIM &   NIMA \\
\hline
Input                                 &  17.42    &  0.8037    &  4.46   \\
CycleGAN~\cite{zhu2017unpaired}       &  20.72    &  0.7825    &  4.37    \\
Exposure~\cite{hu2018exposure}       &  19.74    &  0.8442    &  4.62    \\
EnlightenGAN~\cite{jiang2019enlightengan} &  16.96 &  0.7562   &  4.25    \\
DPE~\cite{chen2018deep}               &  22.36    &  0.8674    &  4.54    \\
Ours  &  \textbf{22.88}    &  \textbf{0.8882}      &  \textbf{4.76}    \\
\hline
\hline
\end{tabular}
\end{table}

\begin{figure*}[htp]
\centering
\subfloat[][Input]{\includegraphics[width=0.245\textwidth]{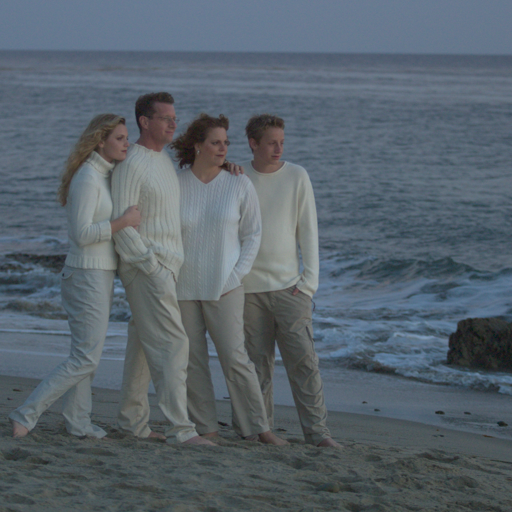}}\hfil%
\subfloat[][CycleGAN~\cite{zhu2017unpaired}]{\includegraphics[width=0.245\textwidth]{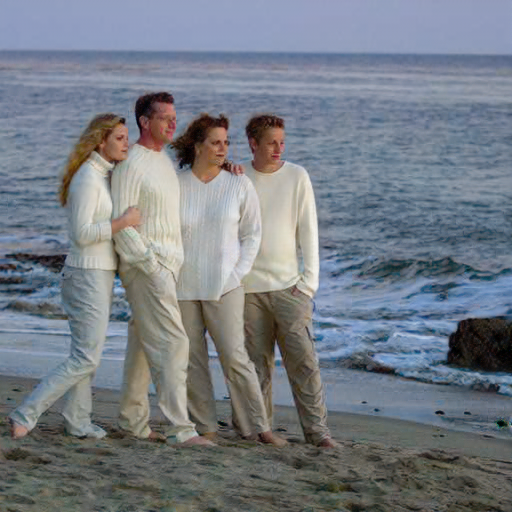}}\hfil%
\subfloat[][Exposure~\cite{hu2018exposure}]{\includegraphics[width=0.245\textwidth]{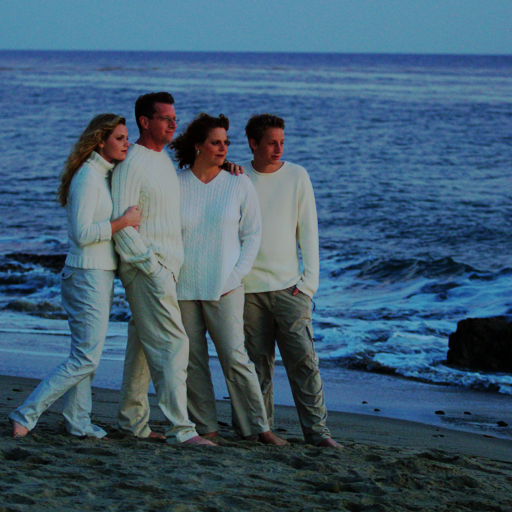}}\hfil%
\subfloat[][EnlightenGAN~\cite{jiang2019enlightengan}]{\includegraphics[width=0.245\textwidth]{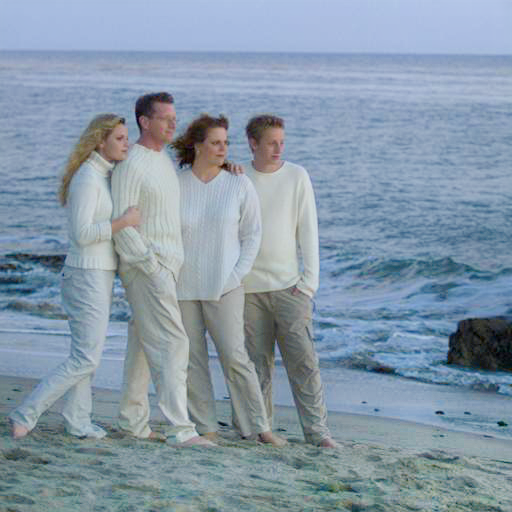}}\\
\vspace{-0.8em}
\subfloat[][DPE~\cite{chen2018deep}]{\includegraphics[width=0.245\textwidth]{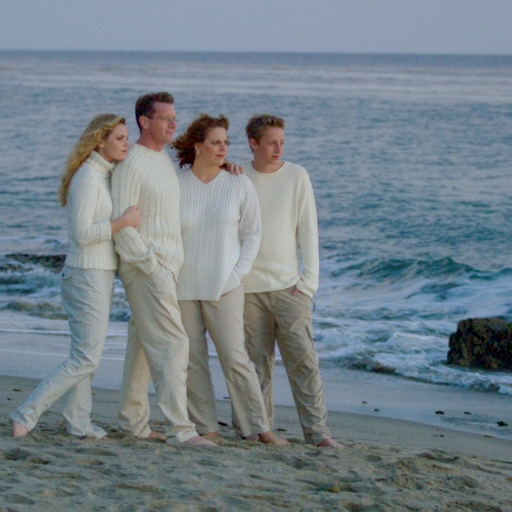}}\hfil%
\subfloat[][Ours (FiveK)]{\includegraphics[width=0.245\textwidth]{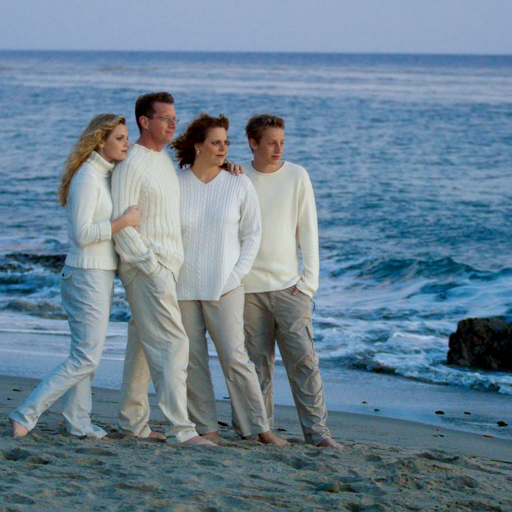}}\hfil%
\subfloat[][Ours (Flickr)]{\includegraphics[width=0.245\textwidth]{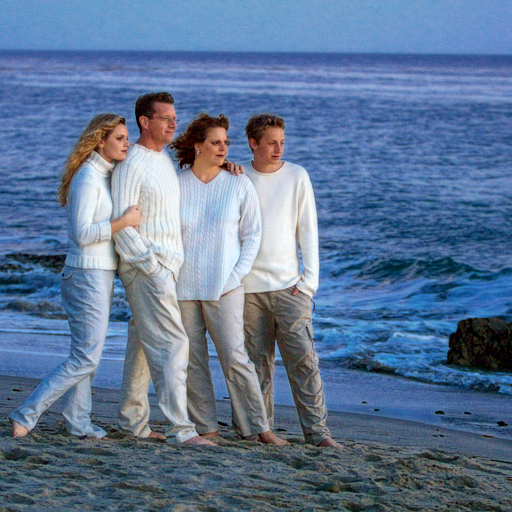}}\hfil%
\subfloat[][Expert-retouched]{\includegraphics[width=0.245\textwidth]{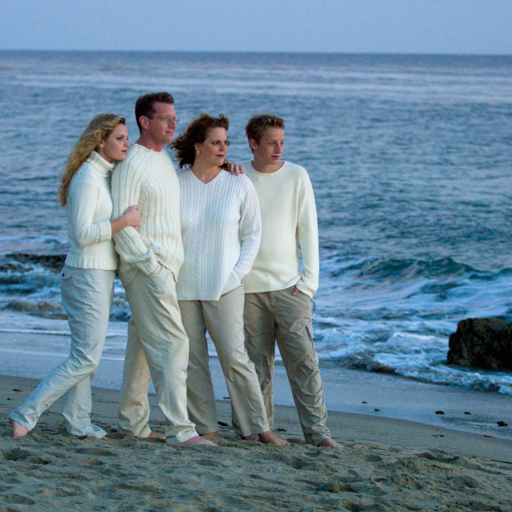}}\\
\caption{Visual quality comparison with state-of-the-art methods (\textit{i.e.}, CycleGAN, DPE, EnlightenGAN, and Exposure) on a test image from the MIT-Adobe FiveK~\cite{bychkovsky2011learning} dataset.}
\label{fig:comparison4}
\end{figure*}

\subsection{Quantitative Comparison}
\label{ssec:quantitative}

Most previous methods for automatic photo quality enhancement are based on supervised learning that requires paired data~\cite{gharbi2017deep, ignatov2017dslr, ren2019low, wang2019underexposed, bychkovsky2011learning}. Recently, a series of works based on GANs or reinforcement learning (RL) attempted to use only unpaired data to solve this tasks. We compared our proposed method with CycleGAN~\cite{zhu2017unpaired}, and three unpaired photo enhancement methods: Deep Photo Enhancer (DPE)~\cite{chen2018deep}, EnlightenGAN~\cite{jiang2019enlightengan} and Exposure~\cite{hu2018exposure}. CycleGAN, DPE, and EnlightenGAN are GAN-based methods and Exposure is an RL and filter-based method.

Table~\ref{tab:quantitative} lists the quantitative comparison results of various models on MIT-Adobe FiveK dataset~\cite{bychkovsky2011learning}. In this table, the best performance of each evaluation metric (\textit{i.e.}, PSNR, SSIM, and NIMA) is boldfaced in black. Please note that the program codes of all models under comparison are downloaded from the link provided by the corresponding authors. Specifically, we used the codes provided by the corresponding authors to retrain the CycleGAN and EnlightenGAN on MIT-Adobe FiveK dataset. We test the Exposure and DPE using the models pre-trained on MIT-Adobe FiveK dataset provided by the corresponding authors, because it achieved better performance than our retrained model. Besides, the Flickr dataset we collected has no ground truth, thus we can only perform qualitative experiments on it. From Table~\ref{tab:quantitative}, one can observe that our proposed UEGAN achieves the best performance in terms of PSNR, SSIM, and NIMA compared with other state-of-the-art image quality enhancement methods trained with unpaired data.

From the experimental results listed in Table~\ref{tab:quantitative}, the following conclusions can be drawn. 1) Our proposed UEGAN, DPE, and Exposure are ranked in the top three in the quantitative comparison and are superior to inputs on all three evaluation metrics. Specifically, our proposed UEGAN has consistently achieved the best performance. 2) Compared with the input, CycleGAN has been obtained worse performance in SSIM and NIMA, which is mainly due to the existence of blocking artifacts in the generated results. 3) Similar to CycleGAN, EnlightenGAN even performed worse on all three evaluation metrics than the input, which may be caused by significant changes in contrast.

\subsection{Qualitative Comparison}
\label{ssec:qualitative}
Besides the superiority in quantitative evaluation, our proposed UEGAN method is also superior to other enhancement methods in qualitative comparison. As shown in Fig.~\ref{fig:comparison1}-\ref{fig:comparison4}, four representative test images were selected from MIT-Adobe FiveK dataset for conducting visual comparisons. One can observe that the input images are diverse and challenging, including: 1) Fig.~\ref{fig:comparison1} (a) is an outdoor scene with normal lighting condition; 2) Fig.~\ref{fig:comparison2} (a) is a landscape image with under-exposed lake surface and buildings; 3) Fig.~\ref{fig:comparison3} (a) is a sky scene with a tiny airplane; 4) Fig.~\ref{fig:comparison4} (a) is a globally under-exposed outdoor scene with little portrait details. Compared to their respective expert retouched versions shown in Fig.~\ref{fig:comparison1} (d) - Fig.~\ref{fig:comparison4} (d), all input images have significantly worse visual experiences. \textit{Additional results are provided in the supplementary material.}

As shown, we obtain some interesting insights. First, the proposed UEGAN trained on our collected Flickr dataset shows the best visual quality among all methods as it generates vivid colors and clear textures. Besides, the results of our proposed UEGAN trained on MIT-Adobe FiveK dataset are satisfactory in enhancing the input image. Second, CycleGAN is less effective in generating vivid colors and also leads to blocking artifacts, which degrade the image quality. In contrast, our method generates visually pleasing results with clear details and sharp structures. Third, Exposure is a filter-based method, which tends to produce over-saturated results and falsely remove textures. However, the results of our proposed UEGAN look natural with good color rendition. Fourth, EnlightenGAN significantly changes the contrast but makes the resulting images dull. On the contrary, our proposed UEGAN can generate satisfactory contrast and natural appearance with the appropriate saturation. Last, DPE produces competitive results compared with ours in structure and contrast enhancement, while it may generate unrealistically looking results. In short, our proposed UEGAN generates natural and pleasing results with satisfactory contrast, vibrant colors and clear details, which is superior to the state-of-the-art methods compared and comparable to the corresponding expert-retouched results.

\begin{table}[t]
\footnotesize
\renewcommand{\arraystretch}{1.5}
\caption{The pairwise comparison preference matrix in user study. EG denotes EnlightenGAN.}
\label{tab:userstudy}
\tabcolsep0.05cm
\centering
\begin{tabular}{lcccccccc}
\hline
\hline
\multirow{2}{*}{}  &  \multirow{2}{*}{Input}  &    CycleGAN &   Exposure  &  EG     &   DPE   &    Ours  &  Ours  &   \multirow{2}{*}{Total} \\
    &   &  \cite{zhu2017unpaired} & \cite{chen2018deep}  & \cite{jiang2019enlightengan}  &  \cite{hu2018exposure} & (FiveK)  & (Flickr)  &   \\
\hline
Input           & -  & 387  & 157  & 171  & 78  & 31  & 16 &  840   \\
CycleGAN           & 733  & -  & 185  & 263  & 95  & 74  &  32 & 1382    \\
Exposure          & 963  & 935   &  -  & 692  & 328  &  253  & 141  &  3312     \\
EG            & 949  & 857  & 428  & -  & 223  & 153  &  92 & 2702    \\
DPE          & 1042  & 1025  & 792 & 897  &  - & 401  & 214 & 4371   \\
Ours (FiveK)           & 1089  & 1046  & 867  & 967  & 719  & -  & 327 & 5015   \\
Ours (Flickr)           & 1104  & 1088  & 979  & 1028  & 906  & 793  & - & 5898   \\
\hline
\hline
\end{tabular}
\end{table}

\begin{figure}[h]
  \centering
  \centerline{\includegraphics[width=0.98\linewidth]{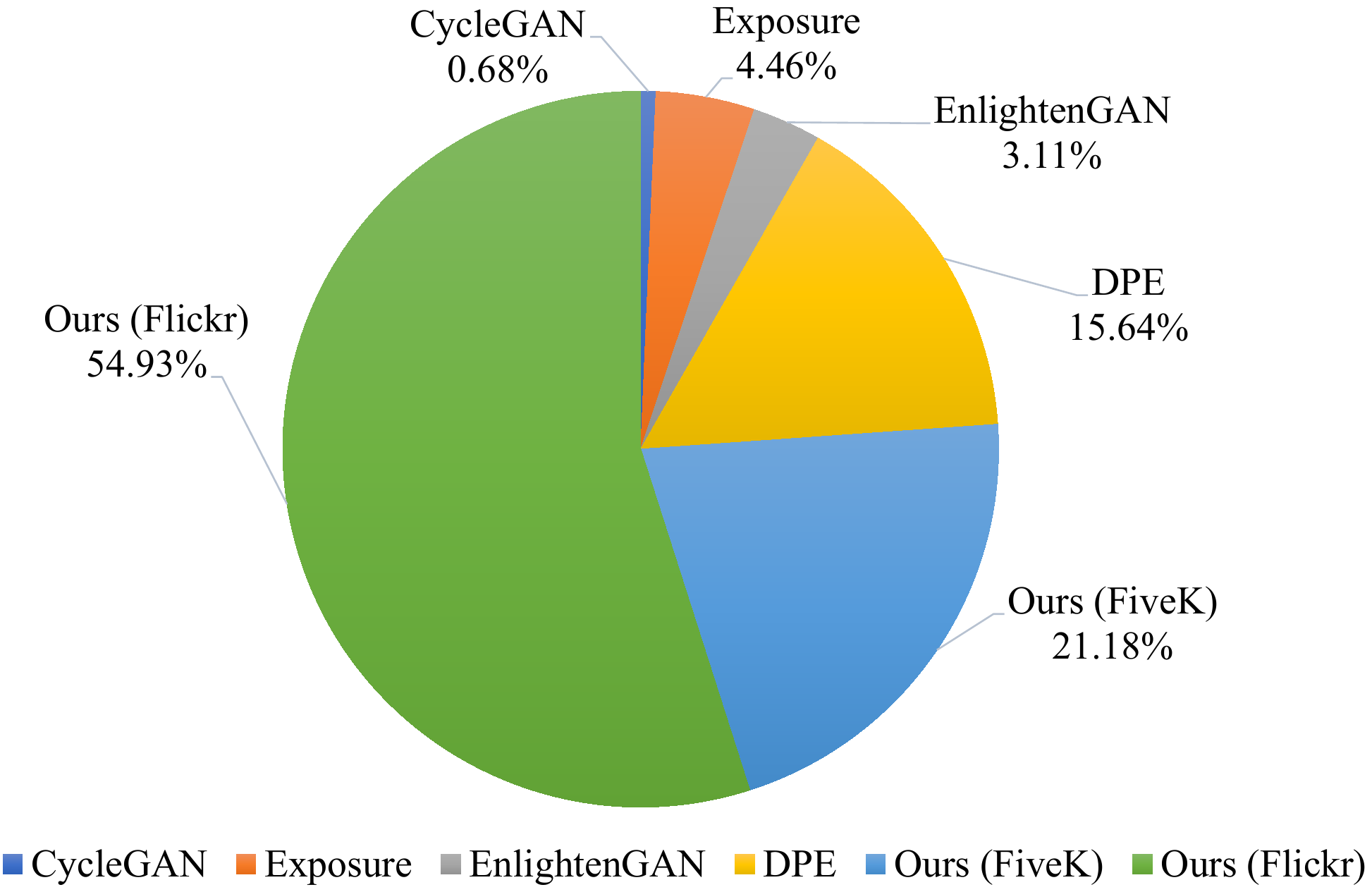}}
  \caption{User preference results of different aesthetic quality enhancement algorithms.}
\label{fig:userstudy}
\end{figure}

\subsection{User Study}
\label{ssec:user}

Our ultimate goal is to learn the implicit characteristics of the target domain to generate high-quality images with similar properties. To measure the subjective quality, we have performed a user study with 28 participants and 40 image sets (\textit{e.g.}, each image set contains 1 test image and the corresponding six generated versions) using pairwise comparisons on six methods (including two versions of our method). The participants are asked to choose his/her favorite result from the displayed pair and the generated images are presented randomly to avoid subjective bias. The corresponding pairwise comparison results are shown in Table~\ref{tab:userstudy}, where each figure indicates the number of times the method in that row outperforms the method in that column. It can be seen that, in all cases, the results of DPE and our proposed UEGAN are preferred much more frequently than the results of other models (\textit{i.e.}, CycleGAN, Exposure, and EnlightenGAN). Among all the comparison methods, the preferred percentages of the proposed UEGAN trained on MIT-Adobe FiveK Dataset~\cite{yan2016automatic} over CycleGAN, Exposure, EnlightenGAN, and DPE are respectively 93.39\%, 77.41\%, 86.34\%, and 64.20\%, and the preferred percentages of our UEGAN trained on our collect Flickr dataset compared with CycleGAN, Exposure, EnlightenGAN, and DPE are 97.14\%, 87.41\%, 91.78\%, and 80.90\%, respectively. It can be seen that the proposed model is selected more frequently than the compared models, which means that the proposed UEGAN can produce more visually pleasing results than all state-of-the-art models in the comparison.

To measure the overall quality, we again randomly selected 100 test images and the corresponding 100 generated results for each model. Each time, six enhanced versions of a test image are present randomly to the participants and asked them to select their favorite one. Finally, 2800 subjective votes are obtained in total and the results are shown in Fig.~\ref{fig:userstudy}. The results show that the enhanced results obtained by our proposed UEGAN are preferred more frequently than those by other methods in the comparison. This further reveals that the proposed UEGAN is superior to all state-of-the-art models in improving the aesthetic quality of the photos.

\begin{figure}[t]
    \centering
    \subfloat[]{
    \begin{minipage}[t]{0.125\textwidth}
        \centering
        \includegraphics[width=0.9\textwidth]{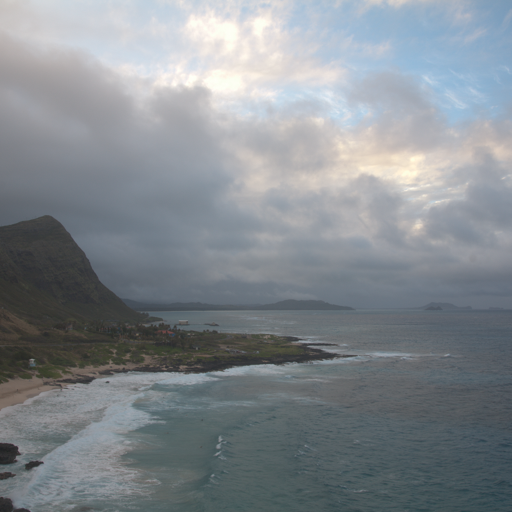} \\ \vspace{0.3em}
        \includegraphics[width=0.9\textwidth]{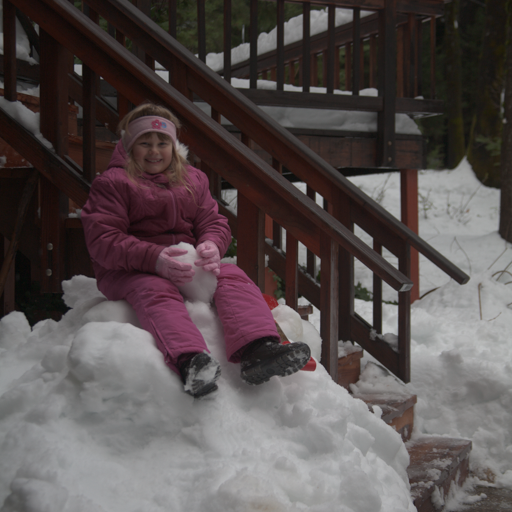}
    \end{minipage}
    }\hspace{-1.5em}
    \subfloat[]{
    \begin{minipage}[t]{0.125\textwidth}
        \centering
        \includegraphics[width=0.9\textwidth]{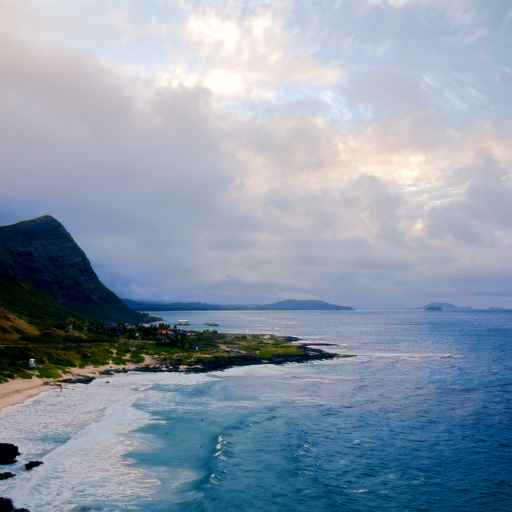}  \\  \vspace{0.3em}
        \includegraphics[width=0.9\textwidth]{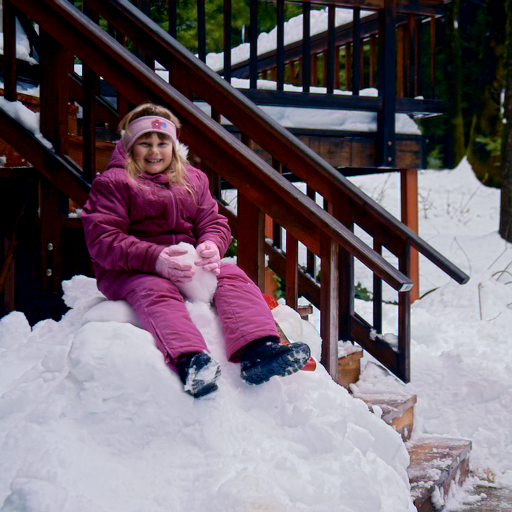}
    \end{minipage}
    }\hspace{-1.5em}
    \subfloat[]{
    \begin{minipage}[t]{0.125\textwidth}
        \centering
        \includegraphics[width=0.9\textwidth]{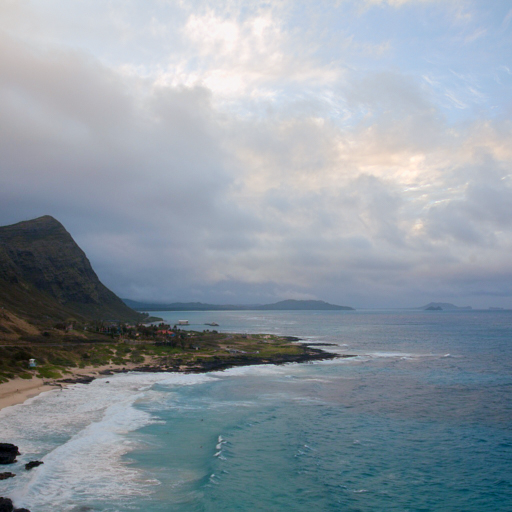} \\  \vspace{0.3em}
        \includegraphics[width=0.9\textwidth]{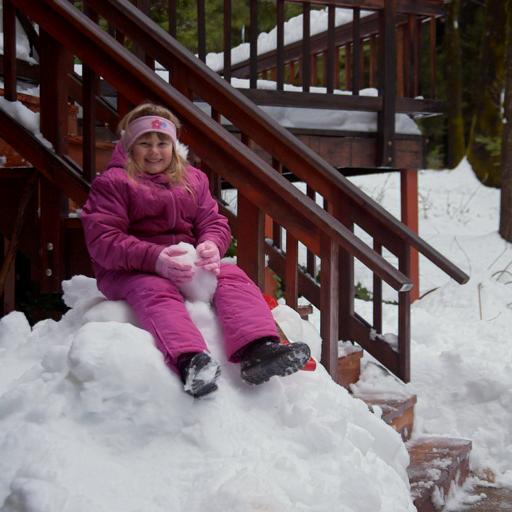}
    \end{minipage}
    }\hspace{-1.5em}
    \subfloat[]{
    \begin{minipage}[t]{0.125\textwidth}
        \centering
        \includegraphics[width=0.9\textwidth]{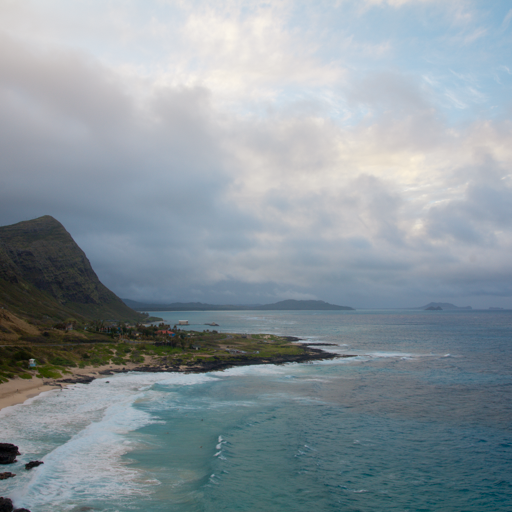} \\  \vspace{0.3em}
        \includegraphics[width=0.9\textwidth]{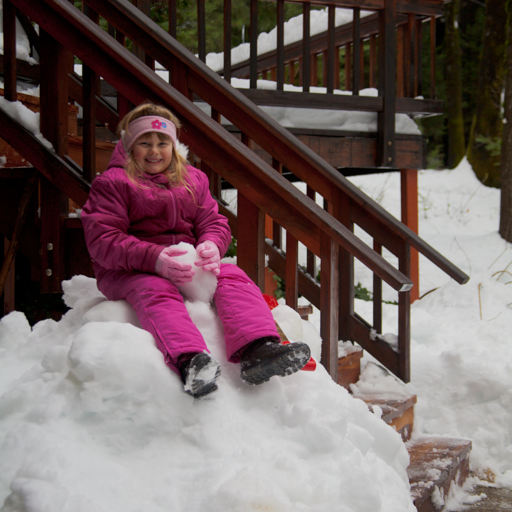}
    \end{minipage}
    }
    \caption{Visual quality comparison results of our proposed UEGAN trained with different loss. (a) Inputs. (b) $\mathcal{L}^G_{\text{qua}}$ + $\mathcal{L}_{\text{fid}}$. (c) $\mathcal{L}^G_{\text{qua}}$ + $\mathcal{L}_{\text{fid}}$ + $\mathcal{L}_{\text{idt}}$. (d) Expert-retouched (\textit{i.e.}, Ground Truth). }
\label{fig:idtloss}
\end{figure}

\section{Analysis and Discussions}
\label{sec:discussion}

\begin{table}[t]
\footnotesize
\renewcommand{\arraystretch}{1.8}
\caption{Average PSNR, SSIM, and NIMA results of enhanced results on MIT-Adobe FiveK Dataset~\cite{bychkovsky2011learning}.}
\label{tab:loss}
\tabcolsep0.25cm
\centering
\begin{tabular}{c|c|c|c}
\hline
\hline
Method  &  PSNR  &    SSIM &   NIMA \\
\hline
Ours w/ $\mathcal{L}^G_{\text{qua}}$,  w/ $\mathcal{L}_{\text{fid}}$, w/o $\mathcal{L}_{\text{idt}}$  &  22.56    &  0.8773      &  4.68    \\
Ours w/ $\mathcal{L}^G_{\text{qua}}$,  w/ $\mathcal{L}_{\text{fid}}$, w/ $\mathcal{L}_{\text{idt}}$  &  \textbf{22.88}    &  \textbf{0.8882}      &  \textbf{4.76}    \\
\hline
\hline
\end{tabular}
\end{table}

\begin{figure}[t]
    \centering
    \subfloat[]{
    \begin{minipage}[t]{0.125\textwidth}
        \centering
        \includegraphics[width=0.9\textwidth]{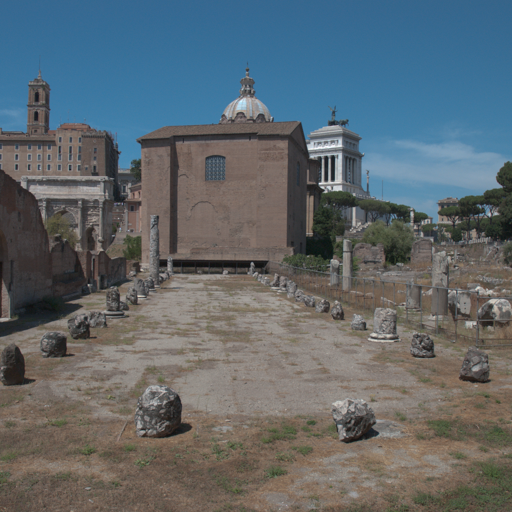} \\ \vspace{0.3em}
        \includegraphics[width=0.9\textwidth]{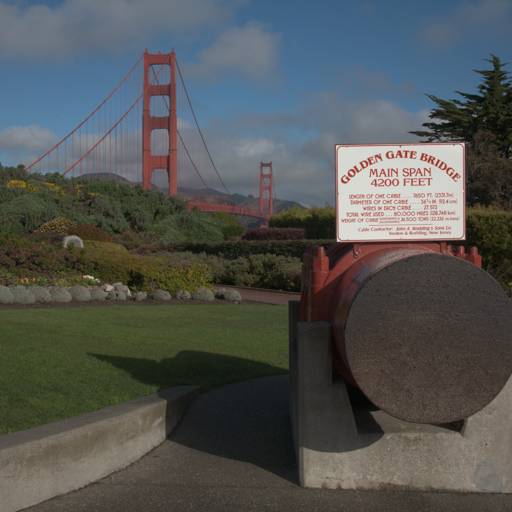}
    \end{minipage}
    }\hspace{-1.5em}
    \subfloat[]{
    \begin{minipage}[t]{0.125\textwidth}
        \centering
        \includegraphics[width=0.9\textwidth]{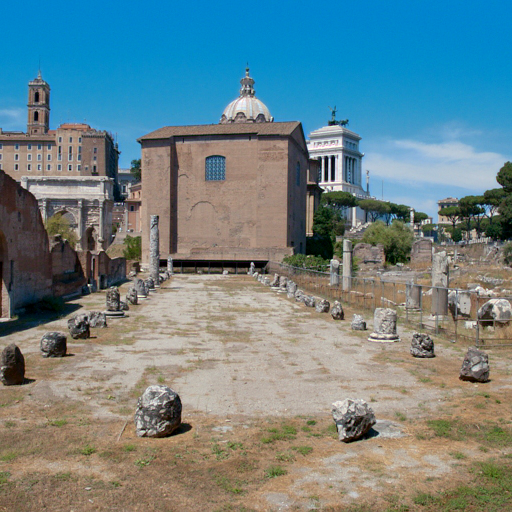}  \\  \vspace{0.3em}
        \includegraphics[width=0.9\textwidth]{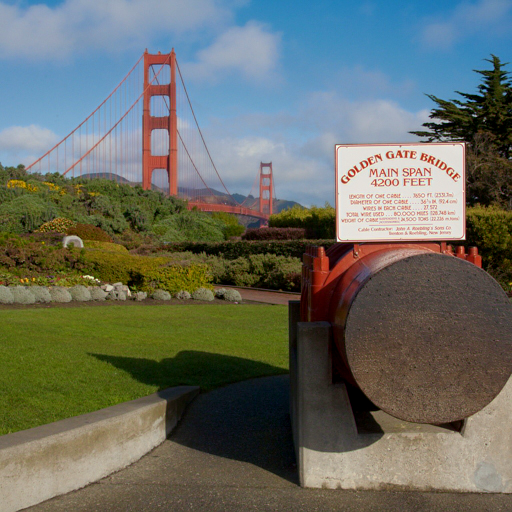}
    \end{minipage}
    }\hspace{-1.5em}
    \subfloat[]{
    \begin{minipage}[t]{0.125\textwidth}
        \centering
        \includegraphics[width=0.9\textwidth]{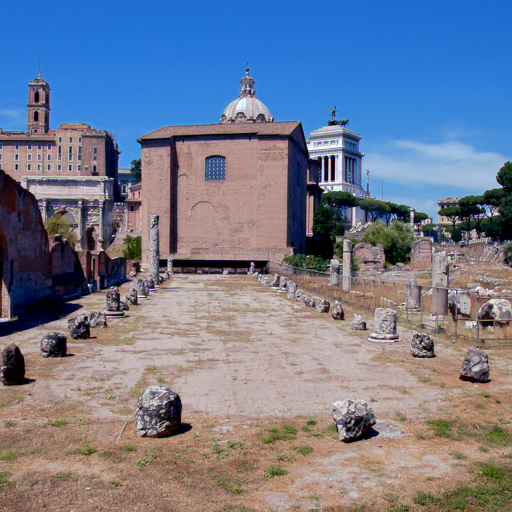} \\  \vspace{0.3em}
        \includegraphics[width=0.9\textwidth]{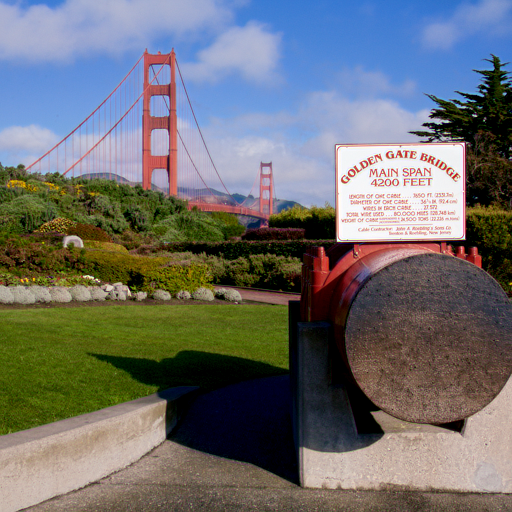}
    \end{minipage}
    }\hspace{-1.5em}
    \subfloat[]{
    \begin{minipage}[t]{0.125\textwidth}
        \centering
        \includegraphics[width=0.9\textwidth]{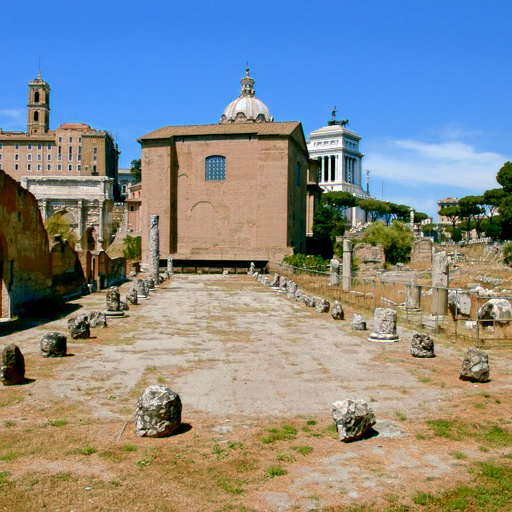} \\  \vspace{0.3em}
        \includegraphics[width=0.9\textwidth]{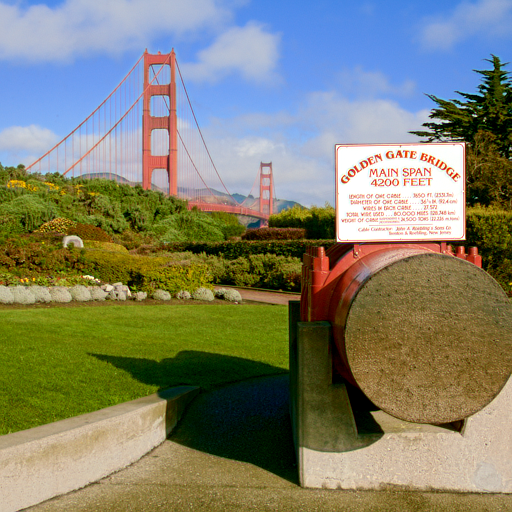}
    \end{minipage}
    }
    \caption{Visual quality comparison results of fidelity loss vs. quality loss. (a) Inputs. (b) - (d) are results obtained by fixing the weighting parameters of $\mathcal{L}_{\text{fid}}$ and $\mathcal{L}_{\text{idt}}$ at 1.0 and 0.1, respectively, and setting $\mathcal{L}^G_{\text{qua}}$ to 0.05, 0.2, and 0.4, respectively.}
\label{fig:advloss}
\end{figure}

\begin{table}[t]
\footnotesize
\renewcommand{\arraystretch}{1.5}
\caption{Comparison of average PSNR, SSIM, and NIMA performance of different network architectures on MIT-Adobe FiveK Dataset~\cite{bychkovsky2011learning}.}
\label{tab:architecture}
\tabcolsep0.28cm
\centering
\begin{tabular}{l|c|c|c}
\hline
\hline
Method  &  PSNR  &    SSIM &   NIMA \\
\hline
GAM + U-Net  &  22.44    &  0.8756      &  4.66    \\
GAM + MM-P  &  17.41    &  0.8037      &  4.46    \\
\hline
UEGAN w/o GAM &  22.54    &  0.8790     &  4.73    \\
UEGAN w/o GAM and MM &  21.87	&   0.8653	&   4.51    \\
UEGAN  &  \textbf{22.88}    &  \textbf{0.8882}      &  \textbf{4.76}    \\
\hline
\hline
\end{tabular}
\end{table}

\begin{figure}[t]
    \centering
    \subfloat[]{
    \begin{minipage}[t]{0.164\textwidth}
        \centering
        \includegraphics[width=0.9\textwidth]{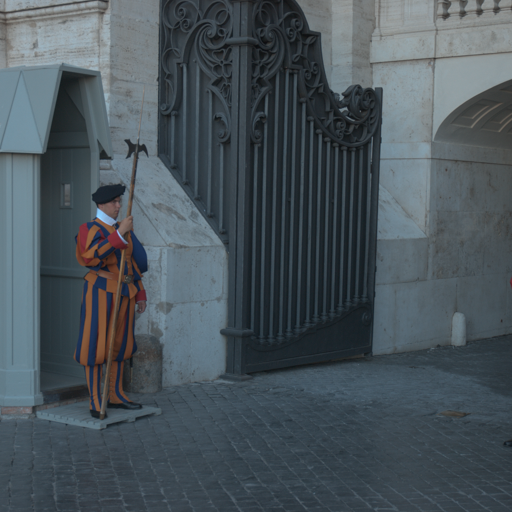} \\ \vspace{0.3em}
        \includegraphics[width=0.9\textwidth]{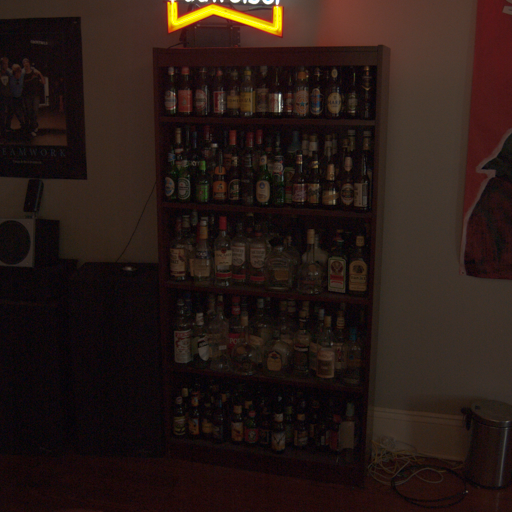}
    \end{minipage}
    }\hspace{-1.6em}
    \subfloat[]{
    \begin{minipage}[t]{0.164\textwidth}
        \centering
        \includegraphics[width=0.9\textwidth]{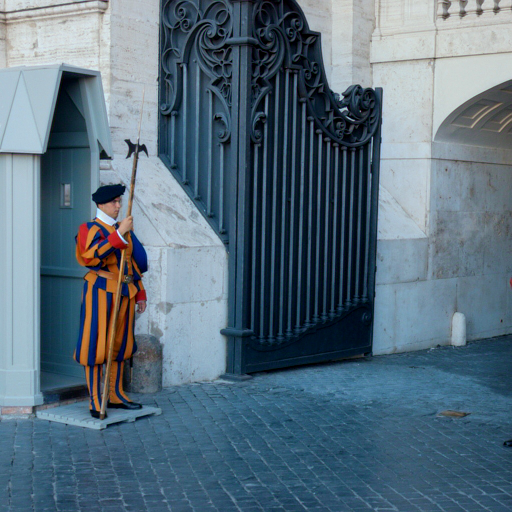}  \\  \vspace{0.3em}
        \includegraphics[width=0.9\textwidth]{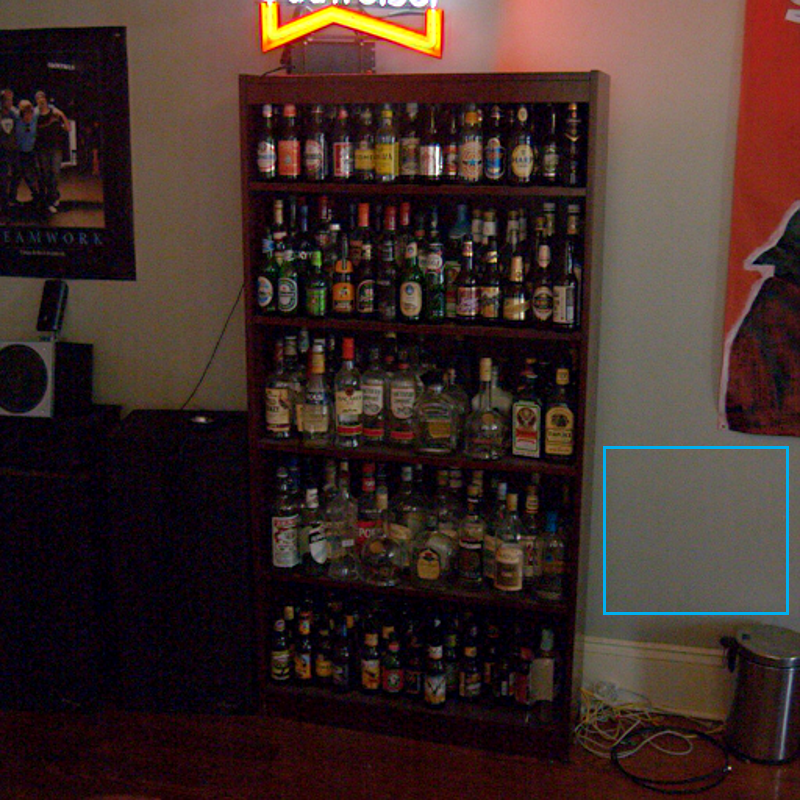}
    \end{minipage}
    }\hspace{-1.6em}
    \subfloat[]{
    \begin{minipage}[t]{0.164\textwidth}
        \centering
        \includegraphics[width=0.9\textwidth]{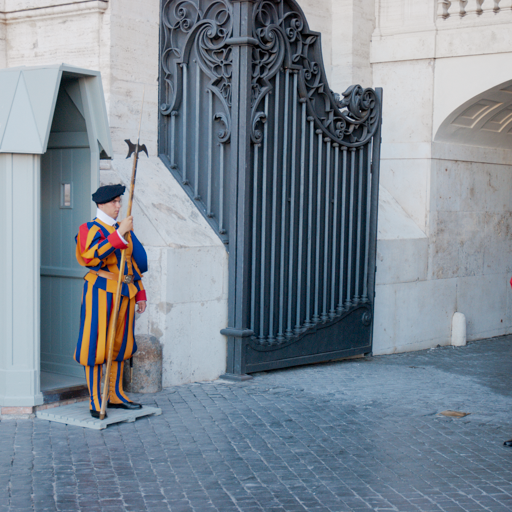} \\  \vspace{0.3em}
        \includegraphics[width=0.9\textwidth]{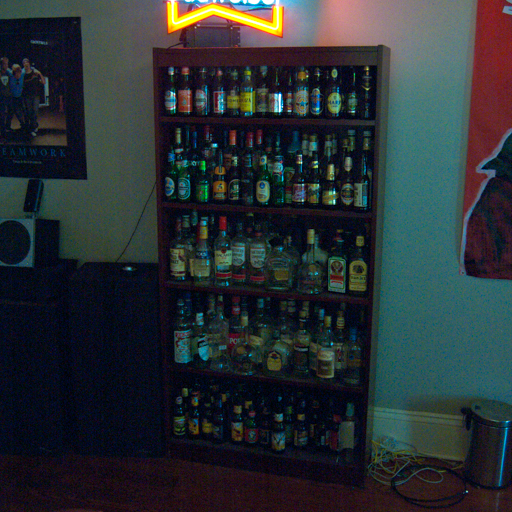}
    \end{minipage}
    }
    \caption{Failure cases generated by our method compared with the ground truth. (a) Inputs. (b) Our results. (c) Ground truth.}
\label{fig:limitation}
\end{figure}

\subsection{Ablation Studies}
\label{ssec:ablation}

\textit{1) Loss analysis:}
In this section, we study the effect of~\textit{quality loss}, ~\textit{fidelity loss}, and~\textit{identity loss} quantitatively and qualitatively. Table~\ref{tab:loss} shows the PSNR, SSIM, and NIMA results achieved by using $\mathcal{L}^G_{\text{qua}}$ + $\mathcal{L}_{\text{fid}}$ and $\mathcal{L}^G_{\text{qua}}$ + $\mathcal{L}_{\text{fid}}$ + $\mathcal{L}_{\text{idt}}$. We can observe that using only $\mathcal{L}^G_{\text{qua}}$ + $\mathcal{L}_{\text{fid}}$ loss achieves better perfomance than the state-of-the-art DPE~\cite{chen2018deep}, EnlightenGAN~\cite{jiang2019enlightengan} and Exposure~\cite{hu2018exposure}, but adding identity loss could further improve quantization performance (\textit{i.e.}, PSNR, SSIM, and NIMA). Fig.~\ref{fig:idtloss} shows two visual comparisons between the results of our proposed UEGAN trained with $\mathcal{L}^G_{\text{qua}}$ + $\mathcal{L}_{\text{fid}}$ and $\mathcal{L}^G_{\text{qua}}$ + $\mathcal{L}_{\text{fid}}$ + $\mathcal{L}_{\text{idt}}$, respectively. It can be observed that the two results generated by our model are more visually pleasing than the input in Fig.~\ref{fig:idtloss} (a). However, compared with the ground truth in Fig.~\ref{fig:idtloss} (d), adding identity loss can suppress over-enhancement to some extent to produce more realistic colors and contrast, as shown in Fig.~\ref{fig:idtloss} (b) and (c).

Fig.~\ref{fig:advloss} shows the results generated by our proposed UEGAN by fixing the weighting parameters of $\mathcal{L}_{\text{fid}}$ and $\mathcal{L}_{\text{idt}}$ at 1.0 and 0.1, respectively, and increasing that of $\mathcal{L}^G_{\text{qua}}$ from 0.05 to 0.4, respectively. We can observe that if we increase the weight of the $\mathcal{L}^G_{\text{qua}}$, the contrast becomes higher and the colors will be more vivid, but the result tends to be over-enhanced and thus loses fidelity. Therefore, we jointly consider fidelity loss, quality loss, and identity loss to improve the visual effect as much as possible while keeping the content the same and avoiding over-enhancement.

\textit{2) Architecture analysis:} In this section, we investigate the effect of each individual component (\textit{i.e.}, global attention module (GAM) and modulation module (MM)) in our proposed UEGAN described in Section~\ref{ssec:generator}. We conduct ablation studies by comparing the proposed UEGAN with the following UEGAN variants: 1) GAM + U-Net: removing the MM and concatenating the features of the first stage of the encoder to those of the penultimate layer; 2) GAM + MM-P: we apply the MM at the pixel level. That is, the generator learns a modulation layer that multiplies the input image with the features of the last layer; 3) UEGAN w/o GAM: removing the GAM from the proposed generator. 4) UEGAN w/o GAM and MM: removing both the GAM and MM from the generator. The quantitative comparison results of all the different architectures are shown in Table~\ref{tab:architecture}. It can be observed that, compared with the traditional U-Net (\textit{i.e.}, GAM+U-Net), our proposed UEGAN achieves the best improvements. Using MM at the feature level can significantly improve the performance than that at the pixel level (\textit{i.e.}, GAM+MM-P). Both GAM or MM lead to better PSNR, SSIM, and NIMA, and combining them can further improve the quantitative performance to achieve the best.

\subsection{Limitations}
\label{ssec:limitations}

The proposed method is completely unsupervised and inevitably has limitations. A typical artifact that is present on the resulting image is color deviation. For example, the color of the ground of the second image in the first row of Fig.~\ref{fig:limitation} is different from that of the input and ground truth. Even though they might produce more pleasing results sometimes coincidentally, this kind of adjustment changes the content and makes the results look unreal. In addition, as shown by the blue box in the second row of Fig.~\ref{fig:limitation}, our method cannot remove noise from the generated results. However, this kind of noise is common in under-exposed images.

\section{Conclusions}
\label{sec:conclusion}

In this paper, we present an~\textit{unsupervised} deep~\textit{generative adversarial network} model developed for image enhancement, call the~\textit{Unsupervised image Enhancement GAN} (UEGAN). The proposed model is able to learn the corresponding~\textit{image-to-image} mapping from a set of images provided by public users with desired characteristics in an~\textit{unsupervised} manner, which makes it possible to learn a~\textit{user-oriented} automatic photo enhancer. We embed the global attention module (GAM) and modulation module (MM) into the generator to capture global features and adjust the features adaptively. In addition, we combine fidelity loss, quality loss, and identity loss with the proposed network to improve the visual quality of the enhanced results. The quantitative and qualitative experimental results show that our proposed method UEGAN is superior to the four state-of-the-art methods.
	
	
	%
	
	
	

	\ifCLASSOPTIONcaptionsoff
	\newpage
	\fi
	
	
	
	%
	
	\bibliographystyle{IEEEtran}
	\bibliography{UEGANRef}

	%

\begin{IEEEbiography}[{\includegraphics[width=1in,height=1.25in,clip,keepaspectratio]{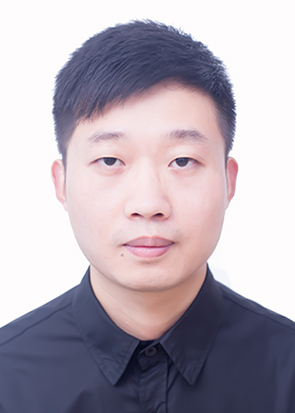}}]
{Zhangkai Ni} (Graduate Student Member, IEEE) received the M.E. degree in communication engineering from the School of Information Science and Engineering, Huaqiao University, Xiamen, China, in 2017. He was a Research Engineer with the School of Electrical and Electronic Engineering, Nanyang Technological University, Singapore, from 2017 to 2018. He is currently a Ph. D. candidate with the Department of Computer Science, City University of Hong Kong, Hong Kong. His current research interests include computer vision, image processing, unsupervised learning, and quality assessment.
\end{IEEEbiography}

\begin{IEEEbiography}[{\includegraphics[width=1in,height=1.25in,clip,keepaspectratio]{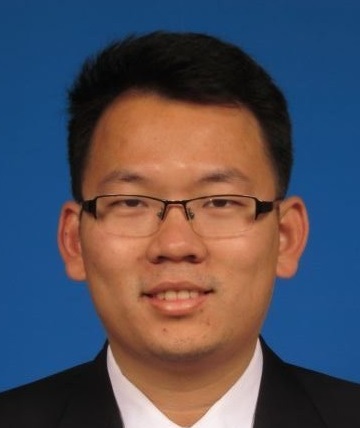}}]
{Wenhan Yang} (Member, IEEE) received the B.S. degree and Ph.D. degree (Hons.) in computer science from Peking University, Beijing, China, in 2012 and 2018. He is currently a postdoctoral research fellow with the Department of Computer Science, City University of Hong Kong. Dr. Yang was a Visiting Scholar with the National University of Singapore, from 2015 to 2016. His current research interests include deep-learning based image processing, bad weather restoration, related applications and theories.
\end{IEEEbiography}

\begin{IEEEbiography}[{\includegraphics[width=1in,height=1.25in,clip,keepaspectratio]{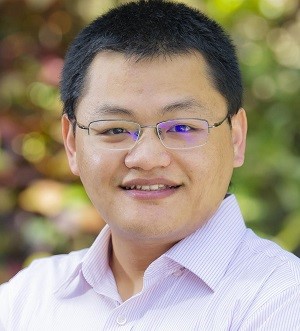}}]{Shiqi Wang} (Member, IEEE) received the B.S. degree in computer science from the Harbin Institute of Technology in 2008, and the Ph.D. degree in computer application technology from the Peking University, in 2014. From Mar. 2014 to Mar. 2016, he was a Postdoc Fellow with the Department of Electrical and Computer Engineering, University of Waterloo, Waterloo, Canada. From Apr. 2016 to Apr. 2017, he was with the Rapid-Rich Object Search Laboratory, Nanyang Technological University, Singapore, as a Research Fellow. He is currently an Assistant Professor with the Department of Computer Science, City University of Hong Kong. He has proposed over 40 technical proposals to ISO/MPEG, ITU-T and AVS standards, and authored/coauthored more than 200 refereed journal/conference papers. He received the Best Paper Award of IEEE Multimedia 2018, the IEEE International Conference on Multimedia and Expo (ICME) 2019, the IEEE International Conference on Visual Communications and Image Processing (VCIP) 2019, the Pacific-Rim Conference on Multimedia (PCM) 2017, and is the coauthor of a paper that received the Best Student Paper Award in IEEE International Conference on Image Processing (ICIP) 2018. His research interests include video compression, image/video quality assessment, and image/video search and analysis.
\end{IEEEbiography}

\begin{IEEEbiography}[{\includegraphics[width=1in,height=1.25in,clip,keepaspectratio]{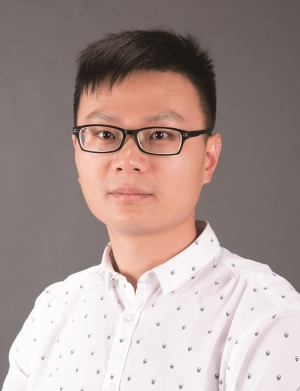}}]{Lin Ma} (Member, IEEE)
received the B.E. and M.E. degrees in computer science from the Harbin Institute of Technology, Harbin, China, in 2006 and 2008, respectively, and the Ph.D. degree from the Department of Electronic Engineering, The Chinese University of Hong Kong, in 2013. He was a Researcher with the Huawei Noah’s Ark Laboratory, Hong Kong, from 2013 to 2016. He was a Principal Researcher with the Tencent AI Laboratory, Shenzhen, China, from 2016 to 2020. He is a currently a Principal Researcher with the Meituan-Dianping Group, Beijing, China. His current research interests lie in the areas of computer vision, multimodal deep learning, specifically for image and language, image/video understanding, and quality assessment.

Dr. Ma received the Best Paper Award from the Pacific-Rim Conference on Multimedia in 2008. He was a recipient of the Microsoft Research Asia Fellowship in 2011. He was a finalist in HKIS Young Scientist Award in engineering science in 2012.
\end{IEEEbiography}

\begin{IEEEbiography}
[ { \includegraphics[width=1in,height=1.25in,clip,keepaspectratio]{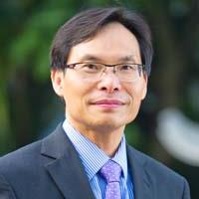} } ]
{Sam Kwong} (Fellow, IEEE) received the B.Sc. degree in electrical engineering from the State University of New York, Buffalo, NY, USA, in 1983 and the M.Sc. degree in electrical engineering from the University of Waterloo, Waterloo, ON, Canada, in 1985, and the Ph.D. degree from the University of Hagen, North Rhine-Westphalia, Germany, in 1996. From 1985 to 1987, he was a Diagnostic Engineer with the Control Data Canada, Mississauga, ON, Canada. He later joined Bell Northern Research Canada, Ottawa, ON, Canada, as a Member of Scientific Staff, and the City University of Hong Kong (CityU), Hong Kong, as a Lecturer with the Department of Electronic Engineering in 1990. He is currently a Chair Professor with the Department of Computer Science, CityU. His research interests include video coding, pattern recognition, and evolutionary algorithms. He is currently the Vice-President of Conferences and Meetings with the IEEE Systems, Man and Cybernetics. He also serves as an Associate Editor of the IEEE TRANSACTIONS ON EVOLUTIONARY COMPUTATION, the IEEE TRANSACTIONS ON INDUSTRIAL ELECTRONICS, and the IEEE TRANSACTIONS ON INDUSTRIAL INFORMATICS, and the Journal of Information Science.
\end{IEEEbiography}

	
	

\end{document}